\documentclass[conference]{IEEEtran}
\usepackage{cite}
\usepackage{amsmath,amssymb,amsfonts}
\usepackage{algorithmic}
\usepackage{graphicx}
\usepackage{textcomp}
\usepackage{xcolor}
\usepackage[ruled, linesnumbered]{algorithm2e}
\usepackage{booktabs} 
\usepackage{multirow}
\usepackage{colortbl}    
\usepackage{graphicx}   
\usepackage{url}
\usepackage{caption}

\def\BibTeX{{\rm B\kern-.05em{\sc i\kern-.025em b}\kern-.08em
    T\kern-.1667em\lower.7ex\hbox{E}\kern-.125emX}}
\begin{document}

\IEEEoverridecommandlockouts

\title{LEAF-SQL: Level-wise Exploration with Adaptive Fine-graining for Text-to-SQL Skeleton Prediction}

\author{
    \IEEEauthorblockN{
        Zhao Tan,
        Xiping Liu\IEEEauthorrefmark{1},
        Qing Shu,
        Qizhi Wan,
        Dexi Liu,
        Changxuan Wan
    }
    \IEEEauthorblockA{
        School of Computing and Artificial Intelligence, Jiangxi University of Finance and Economics, Nanchang, China\\
        tanzhao@stu.jxufe.edu.cn, 
        liuxiping@jxufe.edu.cn, \\
        2202510162@stu.jxufe.edu.cn,
        \{wanqizhi, dexiliu, wanchangxuan\}@jxufe.edu.cn
    }
    \thanks{\IEEEauthorrefmark{1} The corresponding author.}
}

\maketitle

\begin{abstract}
Text-to-SQL translates natural language questions into executable SQL queries, enabling intuitive database access for non-experts. 
While large language models achieve strong performance on Text-to-SQL with prompting, they still struggle with complex queries that involve deeply nested logic or multiple clauses. 
A widely used approach employs SQL skeletons—intermediate representations of query logic—to streamline generation, but existing methods are limited by their reliance on a single structural hypothesis and lack of progressive reasoning.
To overcome these limitations, we propose LEAF-SQL, a novel framework that reframes skeleton prediction as a coarse-to-fine tree search process. LEAF-SQL enables systematic exploration of diverse structural hypotheses with adaptive refinement.
Several key techniques are employed in LEAF-SQL: (1) a three-level skeleton hierarchy to guide the search, (2) a Skeleton Formulation Agent to generate diverse candidates, and (3) a Skeleton Evaluation Agent to efficiently prune the search space.
This integrated design yields skeleton candidates that are both structurally diverse and granularity-adaptive, providing a stronger foundation for the SQL generation.
Extensive experiments show that LEAF-SQL consistently improves the performance of various LLM backbones. On the official hidden test set of the challenging BIRD benchmark, our method achieves 71.6\% execution accuracy, which outperforms leading search-based and skeleton-based methods, affirming its effectiveness for complex queries. 

\end{abstract}

\section{Introduction}
Text-to-SQL, which aims to automatically translate natural language questions into executable SQL queries, is a core technology for intelligent data access. This technology is vital for domains like business intelligence and data analytics because it enables non-expert users to interact with databases using natural language. Recently, Large Language Models (LLMs) have made remarkable progress in this area. By leveraging their strong language capabilities and well-designed prompting, LLMs have achieved state-of-the-art (SOTA) performance on benchmarks like Spider \cite{yu-etal-2018-spider} and BIRD~\cite{li2023bird}. However, LLMs still often struggle with complex tasks involving intricate logic, such as deeply nested queries or complex conditional expressions~\cite{lyu2025sqlo1selfrewardheuristicdynamic, learnat}.

\begin{figure}[t]
    \centering
    \includegraphics[width=1\linewidth]{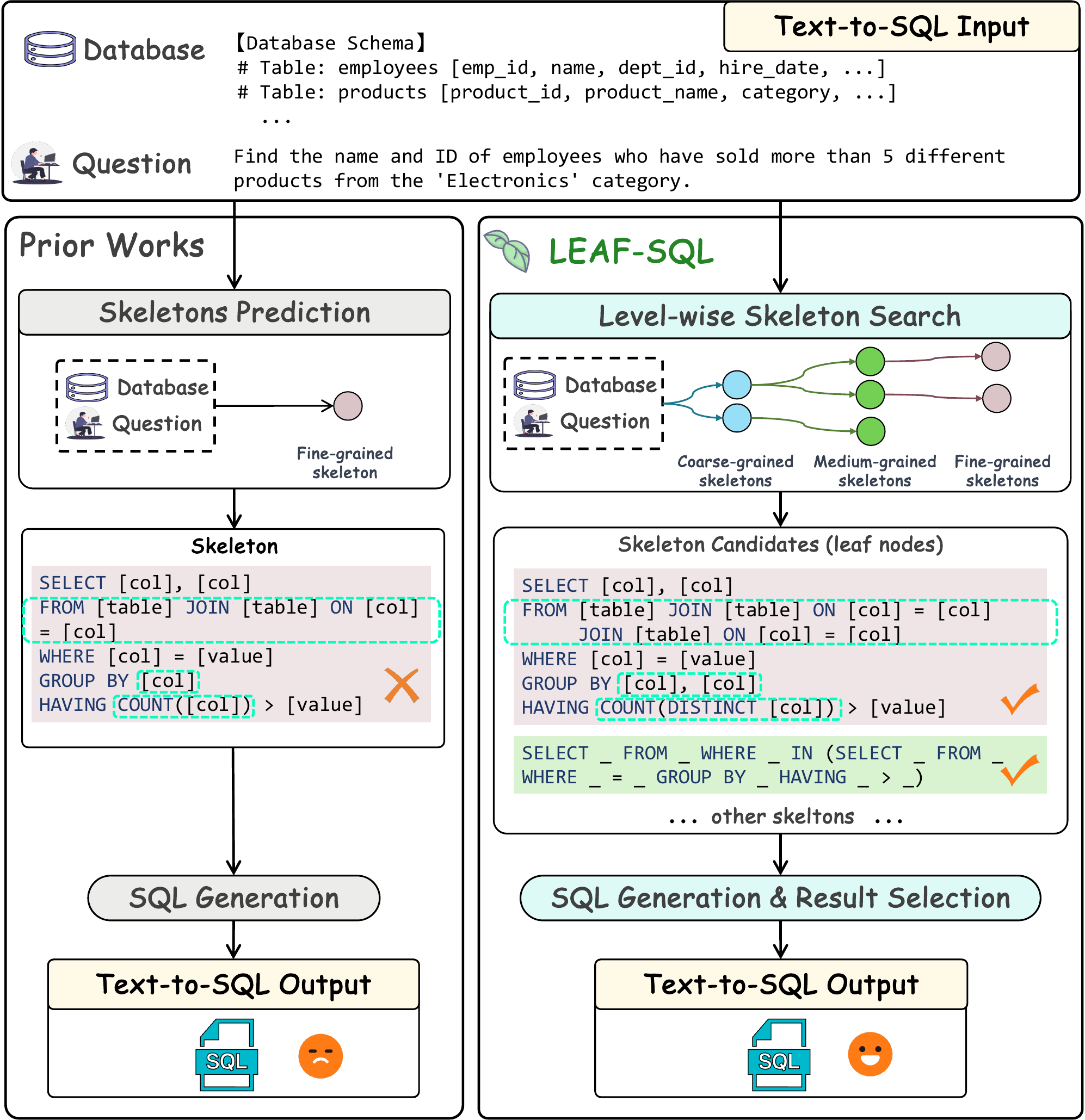}
    \caption{Comparison of LEAF-SQL with prior works. LEAF-SQL enables a search of possible skeletons in a tree-structure space. During this process, branches are evaluated and paths with low confidence are pruned. Finally, LEAF-SQL produces a set of skeleton candidates with different structures and granularities.}
    \label{fig: leaf-sql comparsion}
\end{figure}

To address this challenge, researchers commonly use intermediate representations (IRs) to decompose the complex Text-to-SQL task into more manageable sub-tasks. 
Among these IRs, SQL skeleton has become a particularly popular choice~\cite{10.1145/3589292, li2023resdsql, 10.14778/3681954.3681960, wang2024dacdecomposedautomationcorrection, wu-etal-2025-ucs}.
With SQL skeleton, Text-to-SQL can be split into two steps: first, predicting the skeleton (Text-to-Skeleton), and second, filling the skeleton with schema items like tables and columns (Skeleton-to-SQL).
While this approach has proven effective, existing skeleton-based methods face two critical limitations:
(1) \emph{Limited Exploration of Alternative Structures.} Most methods commit to predicting only one skeleton~\cite{ren2024purplemakinglargelanguage,wang2024dacdecomposedautomationcorrection}. This all-or-nothing strategy is fragile because a natural language question can often be converted into multiple SQL queries with different structures. For example, as shown in Figure~\ref{fig: leaf-sql comparsion}, the question could be answered using either a query with \texttt{JOIN ... GROUP BY} or a nested query with \texttt{WHERE ... IN}. If the single predicted skeleton is flawed, the entire Text-to-SQL process fails with no opportunity for correction.
(2) \emph{Simple Reasoning Mechanism.} Current approaches~\cite{wu-etal-2025-ucs} attempt to generate a relatively detailed skeleton in a single step (a Fine-grained skeleton as depicted in Figure~\ref{fig: leaf-sql comparsion}). Such a design forces the model to resolve all structural complexities simultaneously. This single-pass prediction is prone to error as it lacks a robust mechanism to begin with a high-level logical plan and progressively refine the details.

Inspired by  advanced reasoning techniques like chain-of-thought~\cite{wu-etal-2023-chain} and tree-of-thought~\cite{10.5555/3666122.3666639}, we believe that a tree search paradigm is well-suited to overcome these limitations. Based on this insight, we propose \textbf{LEAF-SQL} (\textbf{L}evel-wise \textbf{E}xploration with \textbf{A}daptive \textbf{F}ine-graining for Text-to-SQL Skeleton Prediction), a novel framework that redefines skeleton prediction as a coarse-to-fine search problem.
LEAF-SQL addresses these limitations in two ways.
First, instead of committing to a single skeleton, LEAF-SQL systematically explores a diverse set of skeletons (see Figure~\ref{fig: leaf-sql comparsion}). 
Second, the search process is inherently progressive and structured. It begins by generating simple, high-level skeleton structures (the ``coarse'' level) and iteratively refines them into more detailed forms (the ``fine'' level). 
Critically, this is not a blind expansion. LEAF-SQL incorporates an evaluation mechanism to intelligently prune unpromising paths and adaptively determine the appropriate level of detail for each candidate. 
This ensures the framework can utilize detailed skeletons when there is high confidence, while safely falling back to more robust, less-detailed ones when uncertainty is present.

\begin{figure}[t]
    \centering
    \includegraphics[width=1\linewidth]{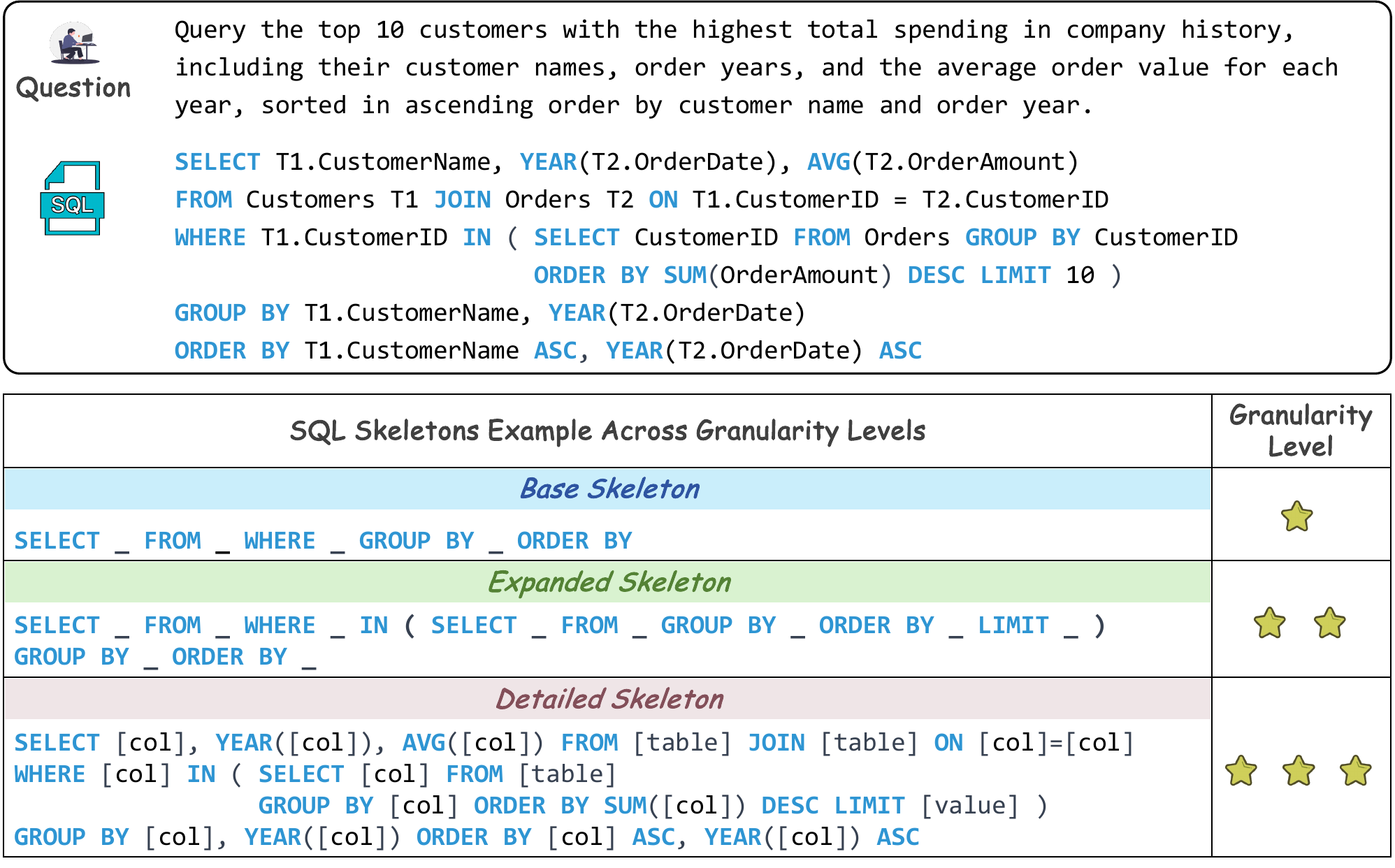}
    \caption{The proposed three-level skeleton hierarchy (Base, Expanded, and Detailed). Base level includes all top level clause nodes; Expanded level incorporates all nested structures; Detailed level specifies details for placeholders and table joins, ensuring a complete, precise representation of the query’s structure.}
    \label{fig: The three-level skeleton hierarchy}
\end{figure}

To realize LEAF-SQL, we introduce several key techniques. 
(1) To structure the coarse-to-fine exploration, we define a \emph{three-level skeleton hierarchy}: Base, Expanded, and Detailed (see Figure~\ref{fig: The three-level skeleton hierarchy}). Each level represents a distinct stage of structural refinement, from a high-level query outline to a detailed logical blueprint. This hierarchy provides the foundational ``levels'' for our level-wise skeleton search.
(2) To drive the forward exploration of the search tree, we design a \emph{Skeleton Formulation Agent} (SkeFor). This agent is responsible for generating new branches by proposing diverse, more detailed skeleton candidates from a parent node. By encouraging structural diversity at each step, SkeFor ensures a broad exploration of the solution space.
(3) To guide the search and ensure its efficiency, we introduce a \emph{Skeleton Evaluation Agent} (SkeEva). Acting as an intelligent quality gatekeeper, SkeEva assesses each newly generated skeleton candidate. It prunes low-quality or incorrect branches, preventing the search from wasting resources on unpromising paths, and enabling the adaptive fine-graining mechanism.
The combination of these components allows LEAF-SQL to produce a set of skeleton candidates that are both structurally diverse and granularity-adaptive, standing out from previous methods (see Figure~\ref{fig: leaf-sql comparsion}). This diverse candidate set offers a stronger, more flexible basis for producing the final SQL query.

The contributions of this paper are summarized as follows:
\begin{itemize}

    \item We propose \emph{LEAF-SQL}, a novel Text-to-SQL framework that reframes skeleton prediction as a coarse-to-fine tree search process. This paradigm systematically explores diverse structural hypotheses while adaptively adjusting the granularity of skeleton candidates.

    \item  We design a set of key technical components to enable this search-based framework: a \emph{three-level skeleton hierarchy} for structured refinement, a \emph{Skeleton Formulation Agent} to generate diverse candidate branches, and a \emph{Skeleton Evaluation Agent} for intelligent pruning of search paths.  

    \item We perform comprehensive experiments to assess LEAF-SQL across multiple Text-to-SQL benchmarks. On the challenging BIRD benchmark, LEAF-SQL achieves 73.5\% execution accuracy on the dev set and 71.6\% on the official hidden test set, outperforming leading existing search-based and skeleton-based methods~\footnote{Our code is available at \url{https://github.com/Atlamtiz/LEAF-SQL}}.

\end{itemize}

\section{Related Works}
\subsection{Text-to-SQL in the Era of LLMs}
\begin{figure}[ht]
\centering
\includegraphics[width=1\linewidth]{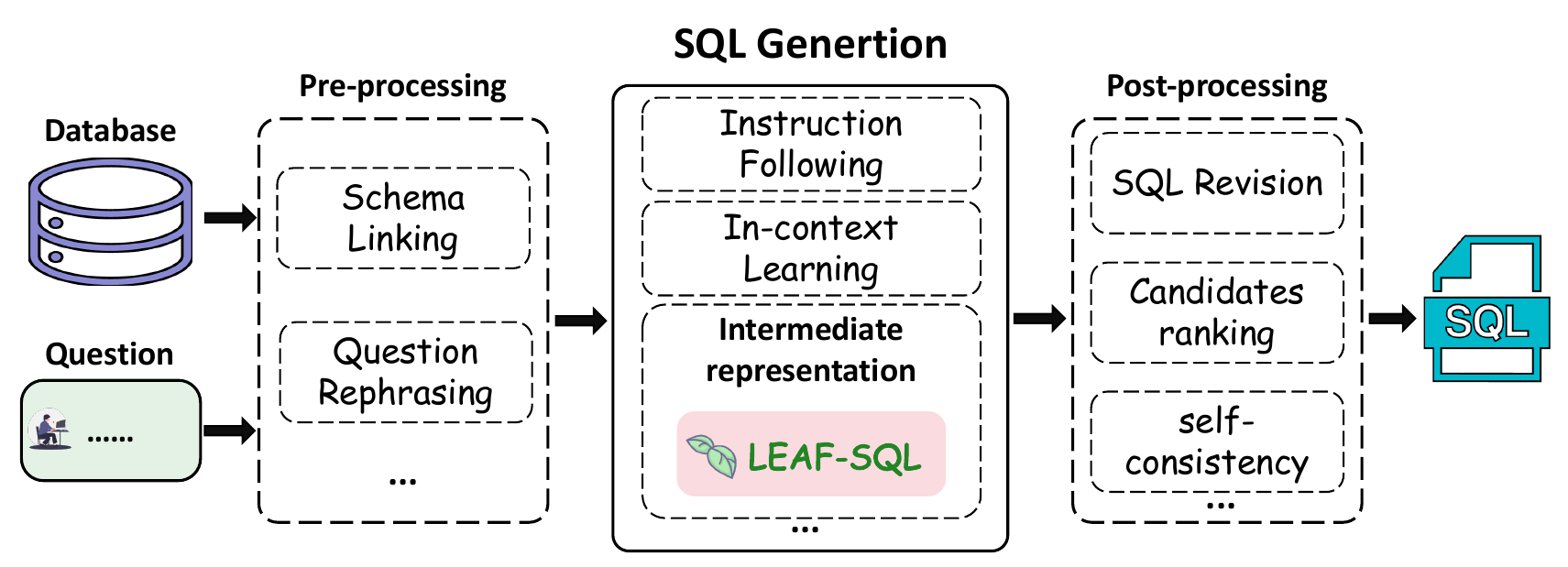}
\caption{The prevailing multi-stage pipeline for Text-to-SQL in the era of LLMs.}
\label{fig:relateworks}
\end{figure}

The Text-to-SQL task aims to automatically translate a natural language question into an executable SQL query, conditioned on a given database schema~\cite{zhong2017seq2sqlgeneratingstructuredqueries, yu-etal-2018-spider, li2023bird, lei2025spider20evaluatinglanguage}. 
Recently, LLMs have achieved strong performance on Text-to-SQL tasks through sophisticated prompting techniques~\cite{11095853}, typically by employing a multi-stage pipeline~\cite{supersql}. As illustrated in Figure~\ref{fig:relateworks}, this pipeline generally consists of three stages: Pre-processing, SQL Generation, and Post-processing.

Pre-processing and post-processing serve as optional but valuable enhancements.
Pre-processing aims to optimize the LLM's input through strategies like schema linking~\cite{rsl-sql}, which identifies task-relevant database elements, and question paraphrasing~\cite{mao-etal-2024-enhancing}, which clarifies the semantics and intent of a natural language question. 
Post-processing refines the output using techniques such as SQL revision~\cite{qu-etal-2025-share, chen2023teachinglargelanguagemodels} or result-driven methods like candidate ranking and self-consistency~\cite{alpha-sql}.

At the heart of this pipeline is the SQL generation stage. To perform this step, existing approaches prompt the LLM using several strategies, including: 
(1) instruction following~\cite{lou-etal-2024-large, tai-etal-2023-exploring}, which guides the model through tailored instructions to perform specific reasoning behaviors (e.g., chain-of-thought prompting); 
(2) in-context learning,  which provides few-shot demonstrations to condition the model on task-relevant examples~\cite{dail-sql}; 
and (3) the use of an intermediate representation. An intermediate representation, such as question decomposition~\cite{mac-sql} or a SQL skeleton~\cite{wu-etal-2025-ucs}, reduces task complexity by breaking the generation process into smaller, more manageable steps.
Our work, LEAF-SQL, operates in the SQL generation stage and adopts SQL skeletons as the intermediate representation.

\subsection{Intermediate Representation with SQL Skeleton}
A SQL Skeleton represents the basic structure of a SQL query, such as \texttt{SELECT \_ FROM \_ WHERE \_}, and is composed of placeholders and SQL keywords. The introduction of skeletons helps to bridge the semantic gap between natural language questions and SQL queries by decomposing the Text-to-SQL task into two sub-tasks: Text-to-Skeleton and Skeleton-to-SQL. For example, UCS-SQL~\cite{wu-etal-2025-ucs} extracts structural information from a question to predict a skeleton, which is then filled with content information to generate the final SQL. Similarly, ZeroNL2SQL~\cite{10.14778/3681954.3681960} uses a smaller language model to generate a skeleton and a more powerful LLM to fill it. RESDSQL~\cite{li2023resdsql} incorporates skeletons into its training phase and proposes a skeleton-aware decoder, which explicitly divides the decoding process into two stages: first generating the skeleton, then the full SQL query. 
Furthermore, the use of skeletons is not limited to intermediate representations. DAIL-SQL~\cite{dail-sql} uses skeletons to retrieve the most relevant demonstrations for a given task, while DAC~\cite{wang2024dacdecomposedautomationcorrection} applies them in a post-processing stage to verify the structural correctness of the generated SQL.

However, these skeleton-based methods typically adopt an ``all-or-nothing'' strategy, predicting a single, fine-grained skeleton in one step. This approach lacks flexibility and is prone to errors on complex queries. In contrast, LEAF-SQL departs from single-step prediction by systematically exploring multiple structural hypotheses through a progressive, coarse-to-fine search.

\subsection{Tree Search Strategy for Text-to-SQL}
Large Language Models are fundamentally conditional probability models, meaning the highest-probability output is not always the correct one. In Text-to-SQL, this ``greedy decoding'' approach often fails to find the optimal solution~\cite{csc-sql}. Consequently, incorporating search-based reasoning frameworks has become a natural direction to explore.

In prior work, both MCTS-SQL~\cite{yuan2025mctssqllightweightllmsmaster} and Alpha-SQL~\cite{alpha-sql} have achieved significant results using tree search. 
MCTS-SQL employs Monte Carlo Tree Search in a post-processing step to correct errors in generated SQL, whereas Alpha-SQL models the Text-to-SQL reasoning process as a sequence of composable actions and then uses a tree search to dynamically find the optimal action path for each given task.

Unlike prior methods, LEAF-SQL integrates tree search directly into the SQL generation stage itself. We reframe skeleton prediction as a search problem, exploring a path from a coarse-grained structure to a fine-grained one.  
By dynamically exploring and refining various skeleton hypotheses, LEAF-SQL provides a more robust and adaptive foundation for Text-to-SQL generation.

\section{Preliminaries}
\label{sec:preliminaries}

\subsection{The Text-to-SQL Task}
The primary objective of the Text-to-SQL task is to translate a natural language question $q$ into an executable SQL query $y$, conditioned on a given database schema $\mathcal{D}$. The schema $\mathcal{D}$ is defined as a collection of tables $\mathcal{D} = \{T_1, T_2, \dots, T_m\}$, where each table $T_i$ consists of a set of columns $T_i = \{c_{i,1}, c_{i,2}, \dots, c_{i,n_i}\}$. The task can be framed as learning a mapping function $f$:
\begin{equation}
    y = f(q, \mathcal{D}).
\end{equation}

\subsection{SQL Skeleton}
To simplify the complex mapping from question to SQL, many methods employ an intermediate representation known as a SQL skeleton~\cite{10.1145/3589292, li2023resdsql, 10.14778/3681954.3681960, wu-etal-2025-ucs}. A SQL skeleton provides a high-level outline of a query’s logical structure, incorporating SQL keywords (e.g., \texttt{SELECT}, \texttt{FROM}) and placeholders (e.g.,  \texttt{\_} or \texttt{[col]}).
However, there is no unified standard for defining skeletons across existing works. This lack of consensus results in highly inconsistent representations: some are overly simplistic, offering poor guidance for generation, while others are excessively detailed, which can easily lead to errors.

To address this inconsistency, we introduce a structured \emph{three-level skeleton hierarchy}: \emph{Base}, \emph{Expanded}, and \emph{Detailed}, as shown in Figure~\ref{fig: The three-level skeleton hierarchy}. Each level offers a distinct degree of detail, progressing from general to specific. We derive these skeletons systematically from a query’s Abstract Syntax Tree (AST), providing a solid foundation for our method (see Section~\ref{sec:The Three-Level Skeleton Hierarchy} for details).

\section{Methodology}
\label{sec:methodology}
In this paper, we introduce LEAF-SQL, a novel skeleton-based Text-to-SQL framework. 
LEAF-SQL redefines the skeleton prediction task as a tree-based search process, generating a set of \emph{structurally diverse} and \emph{granularity-adaptive} skeleton candidates to improve query translation. 
By offering multiple structural options, this approach broadens the model’s decision-making scope, boosting its effectiveness on complex Text-to-SQL tasks.

\begin{figure}[ht]
\centering
\includegraphics[width=1\linewidth]{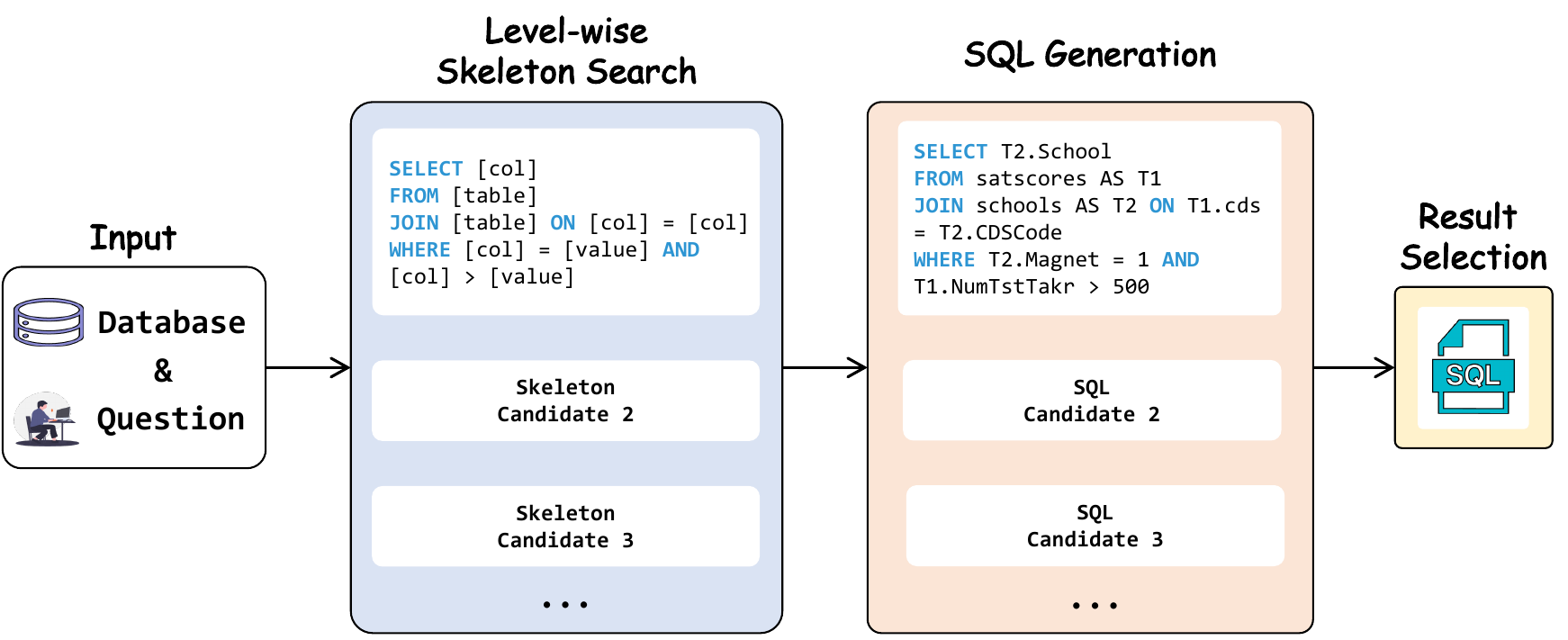}
\caption{Overview of LEAF-SQL.}
\label{fig: Overview of LEAF-SQL}
\end{figure}

As illustrated in Figure~\ref{fig: Overview of LEAF-SQL},  LEAF-SQL is built on three key modules:

\begin{itemize}
    \item \textbf{Level-wise Skeleton Search (Section~\ref{sec:Level-wise Skeleton Search}).} 
    Given a database schema $\mathcal{D}$ and a natural language question $q$, this module aims to produce a skeleton candidate set $\mathcal{S} = \{s_1, \dots, s_n\}$. It introduces an innovative method that transforms skeleton prediction into a flexible, step-by-step tree search, progressing from general to detailed structures.

    \item \textbf{SQL Generation (Section~\ref{sec:SQL_Generation}).} Given the skeleton candidate set $\mathcal{S}$, this module fills each skeleton with database schema elements to produce a SQL candidate set $\mathcal{Y} = \{y_1, \dots, y_n\}$, ensuring accurate query formulation.  

    \item \textbf{Result Selection (Section~\ref{sec:Final_Result_Selection})}. The final SQL is selected from $\mathcal{Y}$ based on a result-driven voting process, refining the output for reliability.
\end{itemize}

\subsection{Level-wise Skeleton Search}
\label{sec:Level-wise Skeleton Search}

\begin{figure*}[ht]
\centering
\includegraphics[width=1\linewidth]{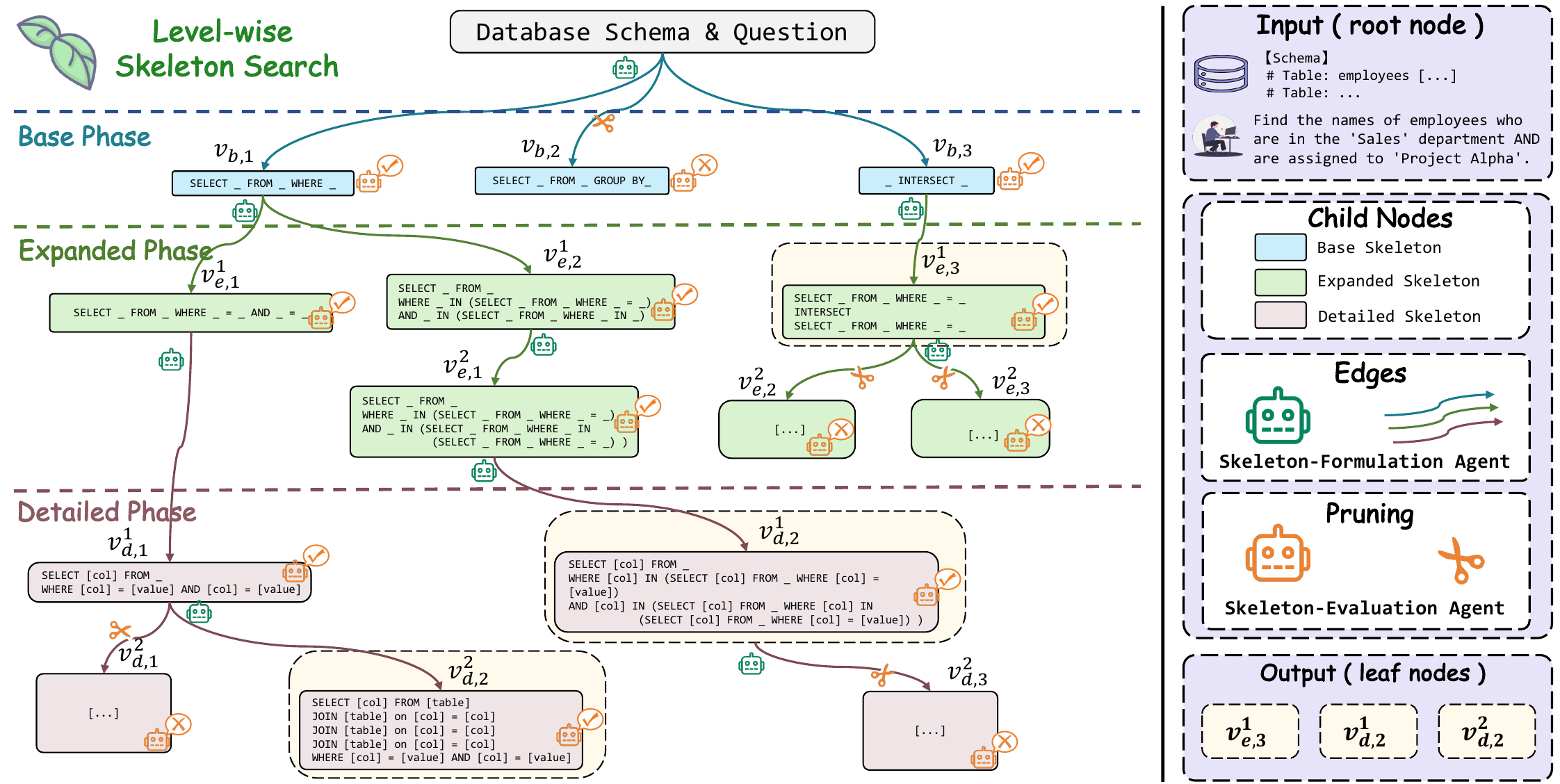}
\caption{An illustration of the Level-wise Skeleton Search. The search has three phases (Base, Expanded, Detailed), each with its own rules, and may include multiple steps.
For node $v_{e,1}^2$: subscript $e$ denotes the exploration phase, superscript $2$ is the step number within the phase, and subscript $1$ is the sibling index.}
\label{fig:method_level_wise_skeleton_search}
\end{figure*}

This module formalizes the skeleton prediction task as a search problem within a structured tree space. Figure~\ref{fig:method_level_wise_skeleton_search} illustrates an example of the search process. The process begins with the root node (i.e., the input pair $(\mathcal{D}, q)$), with child nodes representing skeletons that grow more detailed as the tree deepens.

Each parent node can have multiple child nodes, indicating the derivation of structurally different skeletons from the parent node. Upon completion of the search, the leaf nodes of the search tree are collected to form the structurally diverse and granularity-adaptive skeleton candidate set. This can be formally expressed as:
\begin{equation}
 \mathcal{S} = \{s_1, s_2, \dots, s_n\} = \text{LWSS}(q, \mathcal{D}),
\end{equation}
where $\mathcal{S}$ represents the resulting set of leaf nodes yielded by the Level-wise Skeleton Search (LWSS) process.

This module has several key components:

\begin{itemize}
    \item \textbf{Child Nodes.} Each node corresponds to a specific skeleton, categorized into three levels: \emph{Base level}, \emph{Expanded level}, and \emph{Detailed level}, as detailed in Section~\ref{sec:The Three-Level Skeleton Hierarchy}. This classification enables a gradual, step-by-step exploration across the phases.

    \item \textbf{Edges.} These represent the action of formulating new skeletons, driving the search forward. The process is divided into three phases—\emph{Base phase}, \emph{Expanded phase}, and \emph{Detailed phase}—each with unique rules detailed in Section~\ref{sec:Skeleton Formulation Agent}.

    \item \textbf{Pruning.} This step evaluates the correctness of skeletons to eliminate unpromising branches in the search tree. This process is detailed in Section~\ref{sec:Skeleton Evaluation Agent}.

\end{itemize}

The complete LWSS algorithm is summarized and formalized in Section~\ref{sec:Overall Search Process}.

\subsubsection{The Three-Level Skeleton Hierarchy}
\label{sec:The Three-Level Skeleton Hierarchy}
In the LWSS, each node in the search tree (see Figure~\ref{fig:method_level_wise_skeleton_search}) represents a specific skeleton.  
We categorize these skeletons into three levels based on their granularity: \emph{Base level}, \emph{Expanded level}, and \emph{Detailed level}. These levels are deliberately designed as targets for the corresponding phases of the LWSS process. We label nodes in these levels as $v_b$ (Base level), $v_e$ (Expanded level), and $v_d$ (Detailed level).

\begin{figure}[ht]
    \centering
    \includegraphics[width=0.95\linewidth]{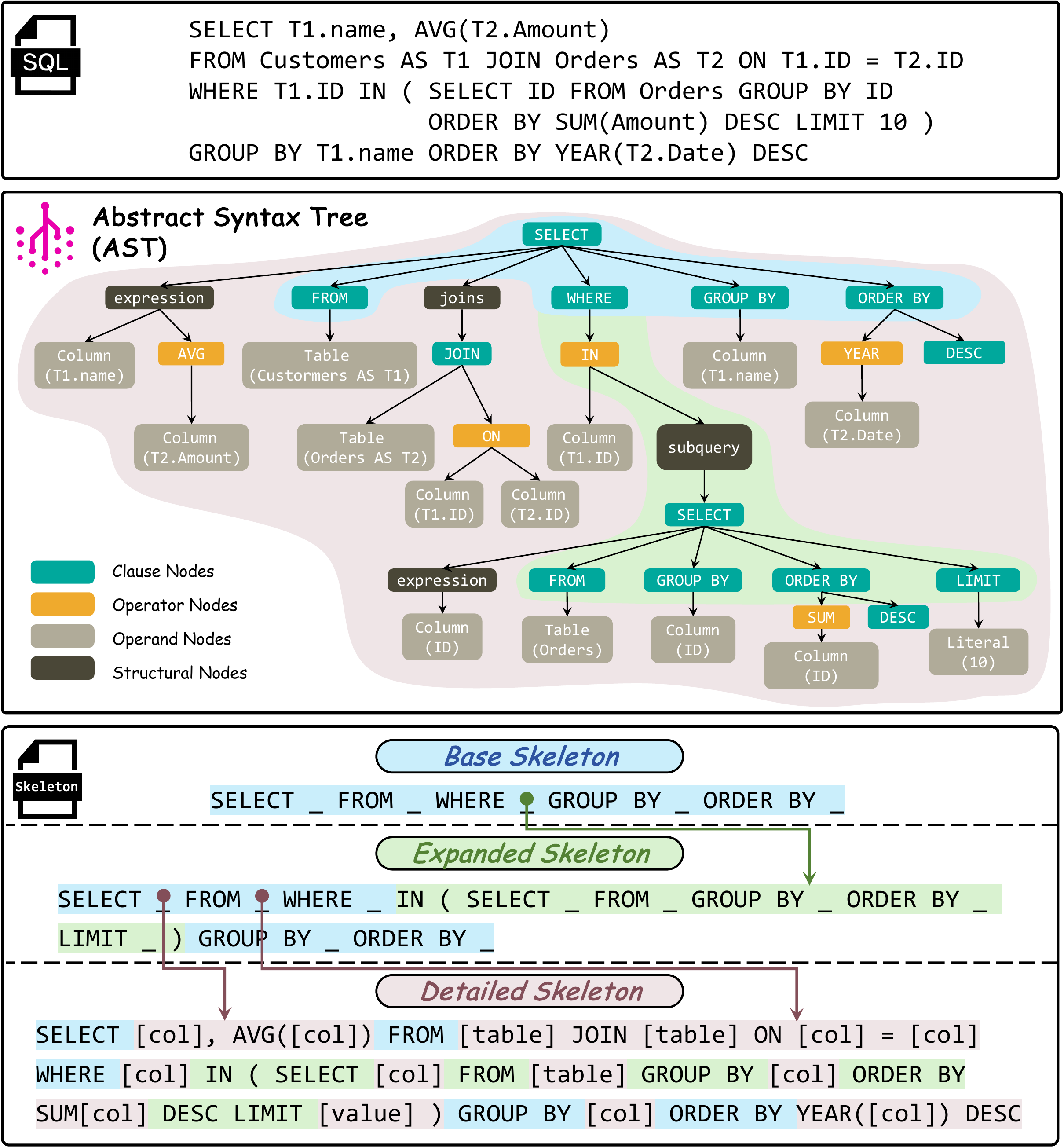}
    \caption{An illustration of how a three-level skeleton hierarchy is derived from a SQL via its Abstract Syntax Tree.}
    \label{fig:method_ast_skeleton_defination}
\end{figure}

The motivation for this hierarchy is to enable progressive, coarse-to-fine reasoning. Instead of predicting an intricate SQL structure in a single high-risk step, we decompose the task into stages, which reduces the burden at each step and creates a multi-layered search space for exploration and validation. The search algorithm first establishes the high-level query structure, then elaborates nested logic, and finally specifies low-level details. Each level serves as a checkpoint; if a path is deemed unpromising, the algorithm can retreat to a simpler parent node and explore alternatives, improving robustness across diverse query complexities.

Specifically, the three skeleton levels are systematically defined using extraction rules applied to the query’s AST, as illustrated in Figure~\ref{fig:method_ast_skeleton_defination}:

\begin{itemize}
    \item \textbf{Base level.} This foundational level includes all top-level clause nodes of the AST (e.g., \texttt{SELECT}, \texttt{FROM}, \texttt{WHERE}). It captures the most fundamental, high-level logical structure of the query.

    \item \textbf{Expanded level.} Building upon the Base skeleton, this level incorporates all nested structures, revealing more complex relationships and nested logic within the SQL query. For example, it might include a subquery structure within a \texttt{WHERE} clause, indicating how conditions depend on additional data sets.

    \item \textbf{Detailed level.} As the most detailed level, this enriches the Expanded skeleton by specifying granular details for placeholders and providing explicit table join details, resulting in a representation that closely mirrors the final SQL structure.
\end{itemize}

\subsubsection{Skeleton Formulation Agent}
\label{sec:Skeleton Formulation Agent}
In LWSS, each edge signifies a step of forward exploration, expanding the search tree by creating more detailed child nodes—specifically, new skeletons—from a parent node.

To implement this crucial step, we introduce the \emph{Skeleton Formulation Agent} (SkeFor). Given an internal node $v$, SkeFor's core function is to produce a set of potential child skeletons to advance the search. This can be formally expressed as:

\begin{equation}
\{ v'_1, \dots, v'_i \} = \text{SkeFor}(\mathcal{D}, q, v),  0 \le i \le m,
\end{equation}
where $v$ is the parent skeleton, $\mathcal{D}$ is the database schema, and $q$ is the question. 
The term $m$ is a predefined hyperparameter representing the maximum number of child nodes (i.e., the branching factor). Note that $m$ defines an upper bound rather than a fixed target; SkeFor is permitted to generate fewer than $m$ nodes.
For the initial Base Phase, which has no parent skeleton, it takes only $(\mathcal{D}, q)$ as input.

A key feature of SkeFor is its \emph{adaptive} strategy, tailored to the three search phases:

\begin{itemize}
    \item \textbf{Base Phase.} This initial phase consists of a single exploration step. SkeFor takes the question and database schema as input and produces a set of diverse Base skeletons, establishing the foundational structures for the search.

    \item \textbf{Expanded Phase.} This phase may involve multiple, iterative exploration steps, depending on the nesting depth of the target SQL query. In each step, SkeFor focuses only on exploring the structure of the next nested level (e.g., nodes $v_{e,2}^1$ to $v_{e,1}^2$ in Figure~\ref{fig:method_level_wise_skeleton_search}). This continues until the deepest nesting is fully explored.

    \item \textbf{Detailed Phase.} This final phase is systematically divided into two steps. The first step explores and specifies the details for all placeholders. The second step then determines the precise table joining information.
    
\end{itemize}

SkeFor is implemented as an LLM-based agent. We guide the LLM to generate the appropriate skeletons for each phase using specifically designed prompts. To ensure structural validity and phase-appropriate granularity, the LLM's output is then normalized by a rule-based script that leverages AST parsing.  

\subsubsection{Skeleton Evaluation Agent}
\label{sec:Skeleton Evaluation Agent}
As the search tree expands, the SkeFor may generate numerous candidate skeletons. Many of these could be incorrect or suboptimal, leading to an inefficient exploration of the vast search space. To ensure search efficiency and focus the process on the most promising paths, a robust pruning mechanism is essential.

\begin{figure}[ht]
    \centering
    \includegraphics[width=1\linewidth]{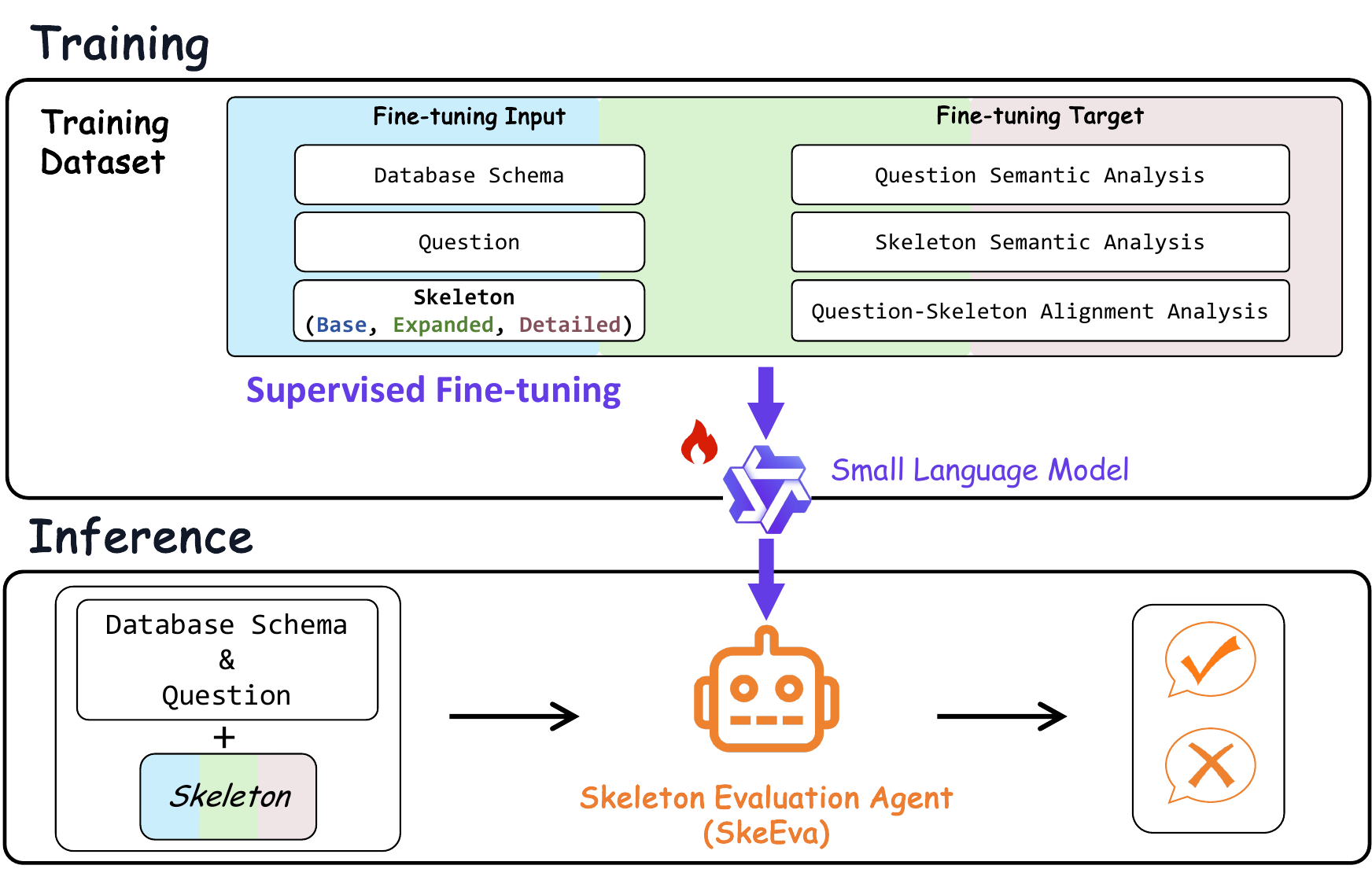}
    \caption{Overall architecture of Skeleton Evaluation Agent training and inference}
    \label{fig:skeleton_evaluation_agent}
\end{figure}

To address this, we introduce the \textit{Skeleton Evaluation Agent} (SkeEva), a specialized component designed to efficiently evaluate the correctness of any given skeleton. We transform a small, open-source language model into this expert evaluator through supervised fine-tuning (SFT). The overall training and inference architecture of SkeEva is illustrated in Figure~\ref{fig:skeleton_evaluation_agent}.

Since skeleton evaluation is a novel task in Text-to-SQL, we constructed a custom SFT dataset. This dataset includes both positive examples derived from ground-truth skeletons and negative examples with realistic errors—such as modified keywords or structural changes—generated programmatically.

Notably, instead of simple labels, each example features a Chain-of-Thought analysis, generated using a powerful LLM (e.g., GPT-4o). This training strategy teaches the model to first analyze the semantics of the question and the skeleton, and then explicitly reason about their alignment. The dataset is also organized by skeleton type—Base, Expanded, Detailed—to tailor the evaluation process.

The model was fine-tuned to minimize the standard autoregressive cross-entropy loss. The target output $Y$ is a Chain-of-Thought analysis comprising three stages: Question Semantic Analysis $A_q$, Skeleton Semantic Analysis $A_{sk}$, and their final Alignment Analysis $A_{align}$. The model learns to generate this structured output based on the database schema $\mathcal{D}$, the question $q$, and the node (i.e., skeleton) $v$. The process can be formalized as generating the concatenated target sequence $A = (A_q, A_{sk}, A_{align})$. The loss function for model parameters $\theta$ is:

\begin{equation}
\label{eq:skeeva_loss}
\mathcal{L}(\theta) = -\sum_{t=1}^{|A|} \log P(a_t | a_{<t}, \mathcal{D}, q, v; \theta),
\end{equation}
where \(a_t\) is the \(t\)-th token in the target sequence $A$.

In summary, SkeEva functions as a highly efficient binary classifier. It takes the database schema, the user's question and a candidate skeleton as input, and outputs a ``True'' or ``False'' verdict. Its role can be formally expressed as:

\begin{equation}
\{\text{True}, \text{False}\} = \text{SkeEva}(\mathcal{D}, q, v).
\end{equation}

\subsubsection{Overall Search Process}
\label{sec:Overall Search Process}

Algorithm~\ref{alg:lwss} details the LWSS algorithm, which constructs the candidate set through a structured \textbf{Generate-Evaluate-Prune} cycle. By coordinating SkeFor and SkeEva, the search progresses dynamically across three phases:

\begin{itemize}
    \item \textbf{Phase 1: Base Initialization.} SkeFor generates diverse Base skeletons from the question. SkeEva immediately validates them, pruning invalid starting points to ensure a robust foundation.

    \item \textbf{Phase 2: Iterative Expansion.} Valid nodes enter a recursive expansion loop. In each step, SkeFor attempts to deepen the logical nesting. This cycle repeats until the generated skeletons no longer increase in nesting depth. This adaptive condition allows the search to accommodate deeply nested queries or skip this phase entirely for flat queries.

    \item \textbf{Phase 3: Detail Finalization.} Leaf nodes from the expansion phase are concretized with placeholders and join conditions. A final SkeEva check ensures correctness before output.
\end{itemize}

\textbf{Termination \& Output.} A search path terminates when a node either (1) achieves the Detailed granularity (e.g., $v_{d,2}^2$ in Figure~\ref{fig:method_level_wise_skeleton_search}) or (2) is a valid node whose potential children are all pruned by SkeEva. The final output $\mathcal{S}$ collects all valid leaf nodes, providing a diverse set of structural hypotheses for SQL generation.

\textbf{Computational Complexity.} To formally analyze the search efficiency, we model the total time cost $T_{\text{total}}$ based on the dynamic search depth $H = L+3$ (where $L$ denotes the query nesting depth) and branching factor $m$:

\begin{align}
    N_{d} &= N_{d-1} \cdot m \cdot (1 - \rho_d), \label{eq:node_count} \\
    T_{\text{total}} &= \sum_{d=1}^{H} N_{d-1} \cdot (T_{gen} + m \cdot T_{eval}), \label{eq:time_cost}
\end{align}

\noindent where $N_d$ represents the surviving nodes at depth $d$, and $T_{gen}, T_{eval}$ denote the unit costs for generation and evaluation. Unlike naive search where $\rho_d \approx 0$ causes exponential growth, LWSS leverages SkeEva to enforce a high rejection rate ($\rho_d \to 1$ for invalid paths), effectively pruning the search space to prevent combinatorial explosion.

\textbf{Implementation Optimization.} To ensure efficiency in practical deployment, our implementation uses an Asynchronous Streaming strategy. Specifically, we decouple the Skeleton Formulation and Evaluation agents via asynchronous queues, enabling pipeline-parallel processing across concurrent queries and improving GPU utilization.

\begin{algorithm}[t]
\small 
\SetAlgoLined
\DontPrintSemicolon
\SetKwInOut{KwIn}{Require}
\SetKwInOut{KwOut}{Ensure}
\SetKwFunction{FRootNode}{RootNode}
\SetKwFunction{FSkeFor}{SkeFor}
\SetKwFunction{FSkeEva}{SkeEva}
\SetKwFunction{FSetParentLinks}{SetParentLinks}
\SetKwFunction{FDepth}{Depth}
\newcommand\mycommfont[1]{\footnotesize\ttfamily\textcolor{blue}{#1}}
\SetCommentSty{mycommfont}

\KwIn{Database schema $\mathcal{D}$, Natural language question $q$}
\KwOut{Candidate skeleton set $\mathcal{S}$}

$\mathcal{T} \leftarrow \{ \FRootNode(\mathcal{D}, q) \}$ \tcp{Initialize tree}

\tcp{\textbf{Phase 1: Base}}
$V_b \leftarrow \{v \in \FSkeFor(\mathcal{D}, q) \mid \FSkeEva(\mathcal{D}, q, v) = \text{True}\}$\;
$\mathcal{T} \leftarrow \mathcal{T} \cup V_b$; \FSetParentLinks{$V_b, \text{root}(\mathcal{T})$}\;

\tcp{\textbf{Phase 2: Expanded}}
$Q \leftarrow V_b$; $V_e^{\text{ready}} \leftarrow \emptyset$\;
\While{$Q$ is not empty}{
    $v_{\text{parent}} \leftarrow Q.\text{dequeue()}$\;
    $V_{\text{ch}} \leftarrow \{v \in \FSkeFor(\mathcal{D}, q, v_{\text{parent}}) \mid \FDepth(v) > \FDepth(v_{\text{parent}})\}$\;
    \eIf{$V_{\text{ch}} = \emptyset$}{
         $V_e^{\text{ready}} \leftarrow V_e^{\text{ready}} \cup \{v_{\text{parent}}\}$\;
    }{
         $V_{\text{ch}} \leftarrow \{v \in V_{\text{ch}} \mid \FSkeEva(\mathcal{D}, q, v) = \text{True}\}$\;
         \If{$V_{\text{ch}} \neq \emptyset$}{
            $Q.\text{enqueue\_all}(V_{\text{ch}})$; $\mathcal{T} \leftarrow \mathcal{T} \cup V_{\text{ch}}$; \FSetParentLinks{$V_{\text{ch}}, v_{\text{parent}}$}\;
        }
    }
}

\tcp{\textbf{Phase 3: Detailed}}
\ForEach{node $v_e \in V_e^{\text{ready}}$}{
    $V_{d}^1 \leftarrow \{v \in \FSkeFor(\mathcal{D}, q, v_e) \mid \FSkeEva(\mathcal{D}, q, v) = \text{True}\}$\;
    $\mathcal{T} \leftarrow \mathcal{T} \cup V_{d}^1$; \FSetParentLinks{$V_{d}^1, v_e$}\;
    \ForEach{node $v_{d} \in V_{d}^1$}{
        $V_{d}^2 \leftarrow \{v \in \FSkeFor(\mathcal{D}, q, v_{d}) \mid \FSkeEva(\mathcal{D}, q, v) = \text{True}\}$\;
        $\mathcal{T} \leftarrow \mathcal{T} \cup V_{d}^2$; \FSetParentLinks{$V_{d}^2, v_{d}$}\;
    }
}

\tcp{Final Step: Extract Leaf Nodes}
$\mathcal{S} \leftarrow \{v \in \mathcal{T} \mid v \in V_{d}^2 \text{ or } v \text{ has no valid children}\}$\;
\Return{$\mathcal{S}$}\;

\caption{Level-wise Skeleton Search (LWSS)}
\label{alg:lwss}
\end{algorithm}

\subsection{SQL Generation}
\label{sec:SQL_Generation}
After the LWSS module provides a candidate set of skeletons $\mathcal{S}$ (i.e., leaf nodes), the SQL Generation module transforms these skeletons into complete, executable SQL queries. This is performed by an LLM with a zero-shot prompt. For each skeleton $s \in \mathcal{S}$, we generate a corresponding SQL query $y$ using the formal process:
\begin{equation}
    y = \text{LLM}(\text{prompt}(\mathcal{D}, q, s)),
\end{equation}
where the $\text{prompt}(\cdot)$ function assembles the database schema $\mathcal{D}$ (structured following the M-Schema~\cite{liu2025xiyansqlnovelmultigeneratorframework} methodology), the question $q$, and the skeleton $s$.

\begin{figure}[ht]
    \centering
    \includegraphics[width=1\linewidth]{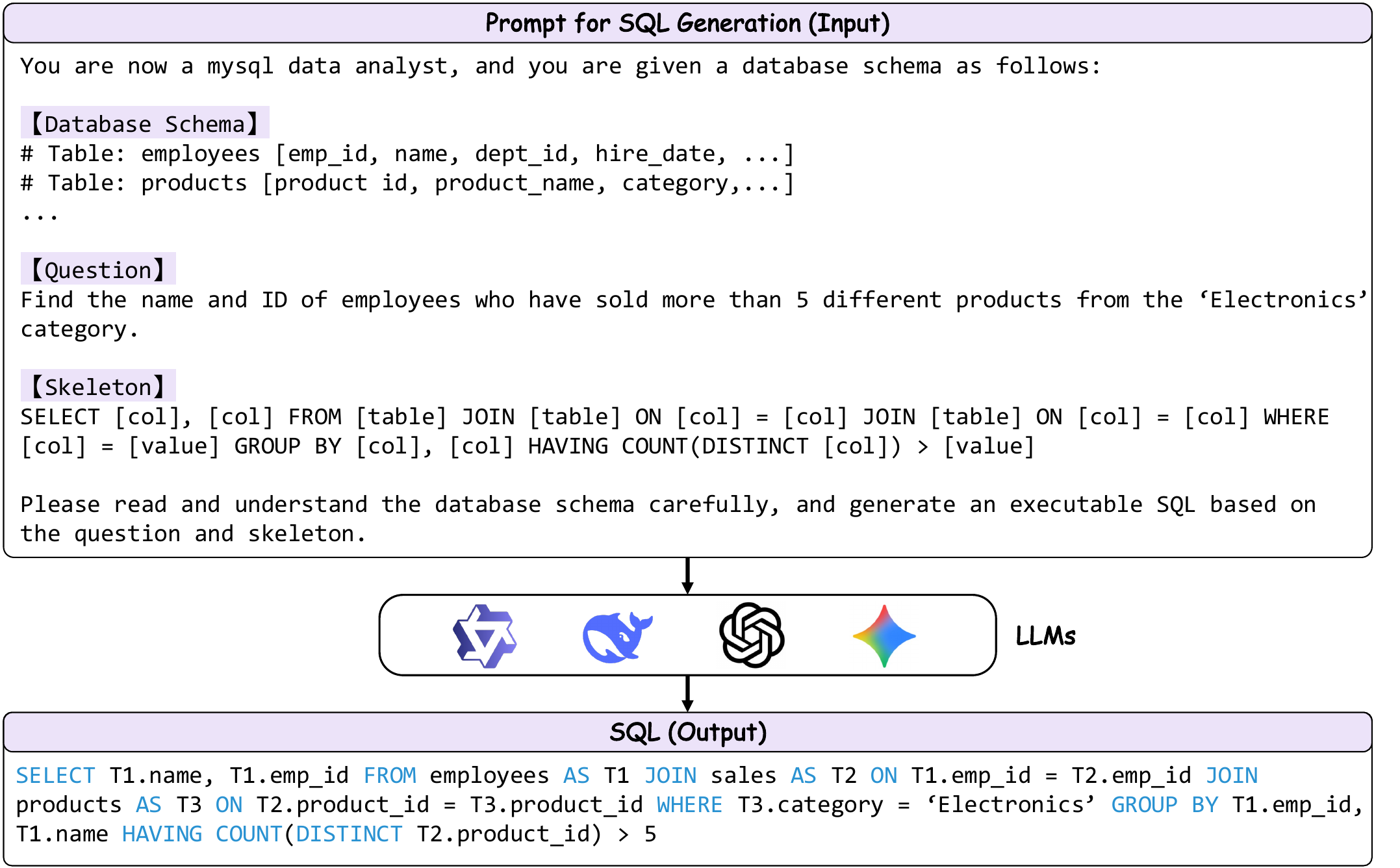}
    \caption{An example of the prompt used for SQL Generation.}
    \label{fig:sql_generation_prompt}
\end{figure}

The skeleton serves as a guiding framework, shaping the LLM’s output to align with the query’s intended structure. An example of this prompt is shown in Figure~\ref{fig:sql_generation_prompt}. The final output of this module is a set of SQL candidates $\mathcal{Y} = \{y_1, \dots, y_n\}$, offering multiple viable interpretations for further refinement.

\subsection{Result Selection}
\label{sec:Final_Result_Selection}
Given the set of generated SQL candidates $\mathcal{Y}$, this module aims to determine the single most reliable query through a result-driven voting process.

First, we execute all queries in $\mathcal{Y}$ and discard any that result in an error or an empty set, yielding a smaller set of valid candidates, $\mathcal{Y}_{\text{valid}}$.
Next, we group these valid queries by their execution results. All queries producing an identical result are placed into the same group, $\{ \mathcal{G}_1, \dots, \mathcal{G}_k \}$. The group containing the most queries is considered the most likely to be correct. We identify the winning group(s) by finding the one(s) with the maximum size, $ N_{\text{max}}$:
\begin{equation}
\label{eq}
N_{\text{max}} = \max_{j \in \{1, \dots, k\}} |\mathcal{G}_j|.
\end{equation}

If there is a single winning group, a query from that group is chosen as the final answer. In case of a tie among multiple groups, we use an LLM as an arbitrator. The LLM is prompted to analyze the distinct results from the tied groups and select the one that best answers the user's question. A query from the LLM-chosen group is then designated as the final answer.

\section{Experiments}

\subsection{Experimental Setup}

\textbf{Benchmarks.} 
Following prior works~\cite{lee-etal-2025-dcg, dai-etal-2025-parsql, yang-etal-2024-synthesizing}, we evaluate our method on five widely-used Text-to-SQL benchmarks. These include two general benchmarks, Spider~\cite{yu-etal-2018-spider} and BIRD~\cite{li2023bird}, as well as three benchmarks designed to assess robustness: SYN~\cite{gan-etal-2021-towards}, REALISTIC~\cite{deng-etal-2021-structure}, and DK~\cite{gan-etal-2021-exploring}. 
Spider is a foundational cross-domain benchmark, containing over 200 databases across 138 different domains. 
BIRD is a more challenging benchmark designed to test real-world scenarios, featuring 95 databases and requiring an understanding of domain-specific knowledge. 
To assess robustness, we also use three variants of the Spider benchmark. SYN tests generalization to paraphrased questions by replacing schema item names with their synonyms. REALISTIC replaces schema mentions in questions with more natural, real-world phrases. DK evaluates the model's ability to reason with domain knowledge not explicitly mentioned in the question.

\textbf{Evaluation Metrics.}
Consistent with prior works, we use Execution Accuracy (EX) as the metric, defined as the percentage of predicted SQL queries that return execution results identical to those of the ground-truth queries.

\textbf{Implementation Details.}
The LWSS module utilizes two key agents. The \textbf{SkeFor} uses Qwen3-14B \cite{yang2025qwen3technicalreport}, guided by phase-specific prompts and a rule-based script leveraging AST parsing to normalize skeleton outputs. In our experiments, we set the search beam width (branching factor) to $m = 3$.
\textbf{SkeEva} is a specialized evaluator created by fine-tuning Qwen3-14B on a custom dataset. This dataset contains 16,000 examples (8,000 positive, 8,000 negative) enriched with Chain-of-Thought reasoning, synthesized via GPT-4o. SkeEva was trained for 5 epochs with a learning rate of 2e-5. 
We apply Demonstration Pruning at inference time by retaining only the demonstration-relevant Gold Schema.
For the \textbf{SQL Generation} and \textbf{Result Selection} modules, we experiment with various powerful LLMs, including models from the GPT~\cite{openai2024gpt4technicalreport}, Gemini~\cite{comanici2025gemini25pushingfrontier}, and Qwen~\cite{qwen2025qwen25technicalreport} series. All performance variations reported in our results tables stem from changing the backbone model in these two modules. For all generative tasks, we set the decoding temperature to 0 for reproducibility.
Main experiments were conducted on a cluster with 8$\times$ NVIDIA RTX 4090 GPUs. We deploy SkeFor and SkeEva as vLLM services with tensor parallelism (TP=4) on GPUs 0--3 and 4--7, respectively. We cap concurrency at 50 and use asynchronous queues to pipeline skeleton generation and evaluation across concurrent queries.

\textbf{Baselines.}
We compare LEAF-SQL against a comprehensive set of baseline methods, categorized as follows: 
(1) Zero-shot: A basic prompt without demonstrations, reflecting the LLM's raw capability. 
(2) DAIL-SQL~\cite{dail-sql}: A classic few-shot method that retrieves and includes similar demonstrations in the prompt. 
(3) MAC-SQL~\cite{mac-sql}: A multi-agent framework that incorporates schema linking, query decomposition, and SQL correction. 
(4) RSL-SQL~\cite{rsl-sql}: A method that introduces a robust schema linking strategy to improve performance.
(5) RESDSQL~\cite{li2023resdsql}: A framework that uses a ranking-enhanced encoder to select relevant schema items and a skeleton-aware decoder to guide the final query generation. 
(6) UCS-SQL~\cite{wu-etal-2025-ucs}: A three-stage Text-to-SQL framework integrating a content pipe for schema linking and a structure pipe for skeleton prediction.
(7) MCTS-SQL~\cite{yuan2025mctssqllightweightllmsmaster}: An approach that employs Monte Carlo Tree Search to iteratively refine the generated SQL query.
(8) Alpha-SQL~\cite{alpha-sql}: A framework that decomposes the Text-to-SQL process into a sequence of actions and uses Monte Carlo Tree Search to find the optimal action combination.

\subsection{Results on General Benchmarks}
To evaluate the overall effectiveness of our approach, we conduct comprehensive experiments on the widely-used Spider and BIRD benchmarks. The main results, including a breakdown by query hardness, are presented in Table~\ref{tab:main experiment}.

\begin{table*}[ht]
    \centering
    \caption{Execution accuracy (EX) on the dev sets of Spider and BIRD.}
    \label{tab:main experiment}
    \resizebox{\textwidth}{!}{
    \begin{tabular}{llccccccccc}
    \toprule 
    \multirow{2}{*}{\textbf{Method}} & \multirow{2}{*}{\textbf{Model}} & \multicolumn{4}{c}{\textbf{BIRD}} & \multicolumn{5}{c}{\textbf{Spider}} \\
    \cmidrule(lr){3-6} \cmidrule(lr){7-11}
     & & \textbf{Simple} & \textbf{Moderate} & \textbf{Challenging} & \textbf{Total} & \textbf{Easy} & \textbf{Medium} & \textbf{Hard} & \textbf{Extra Hard} & \textbf{Total} \\
    \midrule 
    \rowcolor{gray!20} 
    \multicolumn{11}{c}{\textit{\textbf{Prompting-based Methods}}} \\
    \midrule
    \multirow{3}{*}{Zero-shot} 
                    & Qwen2.5-Coder-14B & 61.0 & 42.0 & 31.7 & 52.3 & 86.7 & 77.1 & 63.3 & 51.3 & 72.8 \\
                    & GPT-4o & 63.9 & 46.3 & 38.6 & 56.2 & 94.0 & 80.2 & 70.5 & 57.2 & 78.4 \\
                    & Gemini2.5-pro & 70.3 & 52.8 & 46.9 & 62.8 & 96.0 & 84.2 & 77.3 & 66.3 & 83.3 \\
    \cmidrule(lr){2-11}
    DAIL-SQL & GPT-4 & - & - & - & 59.1 & 91.1 & 88.6 & 75.9 & 62.0 & 86.2 \\
    MAC-SQL  & GPT-4 & 65.7 & 52.7 & 40.3 & 59.4 & - & - & - & - & 86.8 \\
    RSL-SQL  & GPT-4o & 74.4 & 57.1 & 53.8 & 67.2 & - & - & - & - & 87.9 \\
    \midrule
    \rowcolor{gray!20} 
    \multicolumn{11}{c}{\textit{\textbf{Skeleton-based Methods}}} \\
    \midrule
    RESDSQL & T5-3B & - & - & - & - & - & - & - & - & 84.1 \\
    UCS-SQL & GPT-4o & 75.4 & 59.5 & 56.9 & 69.0 & - & - & - & - & 87.3 \\
    \midrule
    \rowcolor{gray!20} 
    \multicolumn{11}{c}{\textit{\textbf{Search-based Methods}}} \\
    \midrule
    \multirow{3}{*}{MCTS-SQL}
                    & Qwen2.5-Coder-7B & 63.0 & 42.2 & 30.3 & 53.6 & - & - & - & - & - \\
                    & GPT-4o & 74.3 & 65.2 & 51.5 & 69.4 & - & - & - & - & 88.7 \\
                    & Gemini2.5-pro & 77.0 & \textbf{69.8} & 56.8 & 72.9 & - & - & - & - & 89.2  \\
    \cmidrule(lr){2-11}
    \multirow{2}{*}{Alpha-SQL}
                    & Qwen2.5-Coder-7B & 72.6 & 59.3 & 53.1 & 66.8 & 94.0 & 89.2 & 76.4 & 63.3 & 84.0 \\
                    & Qwen2.5-Coder-14B & 74.6 & 61.0 & 55.9 & 68.7 & 94.0 & 91.0 & 79.9 & \textbf{72.3} & 87.0 \\
    \midrule
    \rowcolor{gray!20} 
    \multicolumn{11}{c}{\textit{\textbf{Ours}}} \\
    \midrule               
    \multirow{3}{*}{\textbf{LEAF-SQL}}
                    & Qwen2.5-Coder-14B & 75.2 & 58.8 & 57.9 & 67.9 & 94.8 & 90.5 & 80.1 & 68.1 & 86.2 \\
                    & GPT-4o & 77.3 & 61.4 & 60.7 & 70.9 & 96.8 & 92.6 & 83.5 & 69.3 & 88.5 \\
                    & Gemini2.5-pro & \textbf{79.2} & 65.1 & \textbf{64.1} & \textbf{73.5} & \textbf{98.4} & \textbf{93.4} & \textbf{85.8} & 71.7 & \textbf{89.7}  \\
    \bottomrule 
    \end{tabular}
    }
\end{table*}

The results demonstrate the powerful performance of LEAF-SQL. Our framework consistently provides substantial improvements across all backbone models, confirming its general effectiveness. When paired with strong models like Gemini2.5-Pro, LEAF-SQL establishes new SOTA results, achieving 73.5\% EX on BIRD and 89.7\% EX on Spider. This highlights our framework's ability to unlock the full potential of the most capable LLMs.
The primary advantage of LEAF-SQL stems from its unique search-based skeleton prediction. Unlike prior methods, LEAF-SQL operates on a more abstract and efficient level. By exploring a structurally diverse and granularity-adaptive set of high-level logical plans, our framework is better equipped to identify the optimal query structure, an advantage that proves decisive in complex scenarios.

\textbf{Performance on High-Difficulty Queries.}
The advantages of LEAF-SQL are most pronounced on complex tasks. As shown in Table~\ref{tab:main experiment}, our method achieves substantial gains across all difficulty levels, but especially on the most challenging queries. 
On the BIRD-Challenging subset, LEAF-SQL with Gemini2.5-Pro achieves an EX of 64.1\%, which is 7.3 points higher than MCTS-SQL using the same model. This highlights the effectiveness of our approach in complex scenarios. 
Furthermore, skeleton-based methods indeed show a strong aptitude for handling difficult tasks. For instance, UCS-SQL, another skeleton-based method, achieves 56.9\% EX with GPT-4o on the BIRD-Challenging subset, placing it as a strong performer. However, when using the same GPT-4o backbone, our LEAF-SQL reaches 60.7\% EX, demonstrating the superiority of our method even among strong skeleton-based peers. These results affirm that our dynamic, multi-granularity skeleton search is particularly beneficial for tackling the most complex Text-to-SQL challenges.

\subsection{Results on BIRD Test Set}
To ensure a fair and unbiased evaluation, we submitted our code to the official hidden BIRD test set leaderboard\footnote{\url{https://bird-bench.github.io/}}. This blind evaluation process prevents data contamination and reflects true generalization performance. As shown in Table~\ref{tab:bird_test}, LEAF-SQL achieves a SOTA EX of 71.6\% with GPT-4o.
This result surpasses other leading methods, including RSL-SQL which uses the same backbone model, by a significant margin of 2.9 points. It even outperforms Alpha-SQL, a strong method that utilizes tree search. This achievement provides conclusive evidence that our framework's architectural advantages are a key driver of its superior performance on complex, unseen data.

\begin{table}[ht]
\centering
\caption{Execution accuracy (EX) on the hidden BIRD test set. Results are obtained from the official leaderboard.}
\label{tab:bird_test}
\resizebox{0.68\linewidth}{!}{
\begin{tabular}{llc}
\toprule[1pt]
\textbf{Method} & \textbf{Model}  & \textbf{EX} \\
\midrule[0.5pt]
MAC-SQL &  GPT-4  & 59.6 \\
RSL-SQL &  GPT-4o  & 68.7 \\
Alpha-SQL &  Qwen2.5-Coder-32B  & 70.2 \\
\midrule[0.5pt]
LEAF-SQL & GPT-4o  & \textbf{71.6} \\
\bottomrule[1pt]
\end{tabular}
}
\end{table}

\subsection{Results on Robustness Benchmarks}
To assess the generalization capabilities of our method under specific linguistic and knowledge-based challenges, we evaluate LEAF-SQL on three robustness benchmarks: SYN, REALISTIC, and DK. These benchmarks are variants of the Spider benchmark, specifically designed to test a model's resilience to paraphrased questions (SYN), real-world linguistic variations (REALISTIC), and the need for implicit domain knowledge (DK).

As shown in Table~\ref{tab:robust_benchmark}, LEAF-SQL consistently outperforms the compared baselines. When using GPT-4o as the backbone, our method achieves an average accuracy of 79.2\%. By exploring a diverse set of structural skeletons, LEAF-SQL is more resilient to the variations introduced in these specialized benchmarks.

\begin{table}[t]
\centering
\caption{Execution accuracy (EX) on SYN, REALISTIC and DK.}
\label{tab:robust_benchmark}
\resizebox{1\linewidth}{!}{
    \begin{tabular}{llcccc}
    \toprule[1pt]
    \textbf{Methods} &\textbf{Model} & \textbf{SYN} & \textbf{REALISTIC} & \textbf{DK} & \textbf{AVG.}\\
    \midrule[0.5pt]
    Zero-shot & GPT-4o & 69.8 & 72.2 & 66.5 & 69.5    \\
    DAIL-SQL & GPT-4 & 72.2 & 79.5 & 71.2 & 74.3     \\
    MAC-SQL & GPT-4 & 75.2 & 81.0 & 73.6 &  76.6     \\
    
    \midrule[0.5pt]
    LEAF-SQL & GPT-4o & \textbf{76.9} &  \textbf{82.7} &  \textbf{77.9} & \textbf{79.2} \\
    \bottomrule[1pt]
    \end{tabular}
}
\end{table}

\section{Analysis}
In this section, we conduct a series of in-depth analyses to provide a comprehensive understanding of LEAF-SQL from multiple perspectives. 

\subsection{Analysis of Granularity Potential and Realization Difficulty}
\label{sec:Analysis of Granularity Potential and Realization Difficulty}
This subsection quantifies the capabilities of our proposed three-level skeletons by comparing their theoretical potential against their practical realization difficulty. This analysis highlights the core challenge our search-based framework is designed to address.

We conducted two experiments on the BIRD dev set with the GPT-4o model. 
First, to measure theoretical potential, we evaluated the Oracle Skeleton Performance. We programmatically extracted perfect Base, Expanded, and Detailed skeletons from each ground-truth query. These oracle skeletons then guided a GPT-4o model to generate the SQL. 
Second, to assess realization difficulty, we measured the Single-Step Predicted Skeleton Performance. We prompted GPT-4o to directly predict a skeleton for each granularity using a few-shot approach, and these predicted skeletons then guided the same model for SQL generation.

\begin{figure}[ht]
    \centering
    \includegraphics[width=1\linewidth]{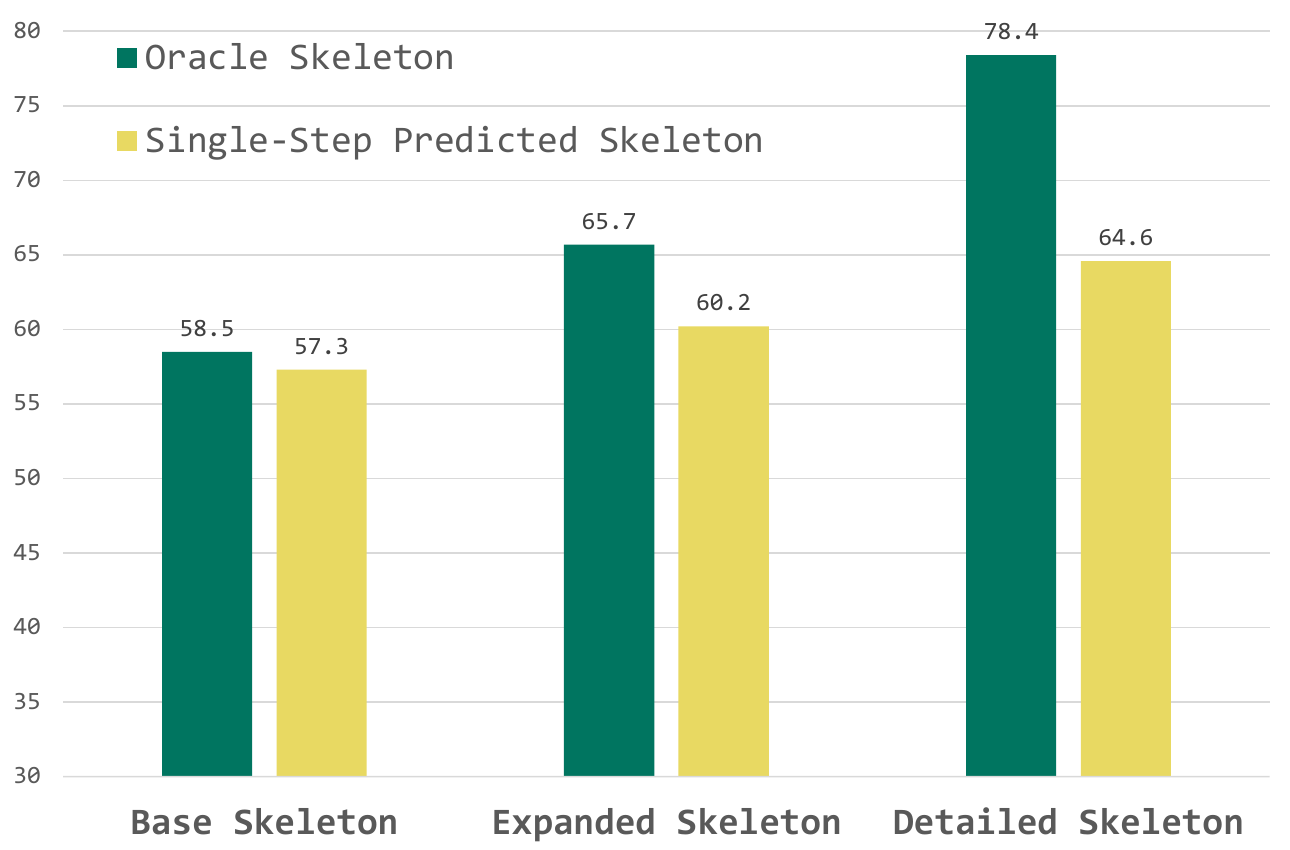}
    \caption{Performance comparison between using Oracle Skeletons and Single-Step Predicted Skeletons on the BIRD dev set. While oracle performance scales with granularity, the gap between oracle and predicted performance widens, highlighting the challenge of single-step prediction for fine-grained skeletons.}
    \label{fig:analysis_gra_level}
\end{figure}

The results in Figure~\ref{fig:analysis_gra_level} reveal a clear trend: oracle performance increases with skeleton granularity, with the Detailed skeleton achieving the highest execution accuracy (78.4\%), making it the most informative representation. 
However, predicted skeleton performance grows more slowly. The Detailed skeleton reaches 64.6\%, trailing its oracle upper bound by 13.8 points. Moreover, the improvement from Expanded (60.2\%) to Detailed (64.6\%) is modest relative to the additional complexity, suggesting that directly predicting fine-grained structures remains challenging.

This ``potential-reality gap'' is the central motivation for LEAF-SQL. Instead of relying on a single prediction step, LEAF-SQL adopts a progressive, search-based strategy that incrementally refines skeletons from coarse to fine while validating intermediate results, thereby improving reliability and better exploiting the potential of fine-grained skeletons.

\subsection{Ablation Study}
To quantify the contribution of each core component within LEAF-SQL, we conducted an ablation study on the BIRD dev set using GPT-4o as the backbone model. 
The results are presented in Table~\ref{tab:ablation}. We compare our full method against three major variants: (1) w/o Diversity (Single-Path Search): We constrain the SkeFor to a single-path search by setting its exploration parameter to $m=1$. (2) w/o Pruning (Brute-force Expansion): This variant disables the SkeEva. All generated candidate skeletons, regardless of their quality, are retained and expanded in subsequent steps. (3) w/o Voting (Random Selection):  This variant disables the final voting-based selection mechanism and instead randomly selects one SQL query from the SQL candidate set as the final result.

\begin{table}[ht]
\centering
\caption{Ablation study of LEAF-SQL on the BIRD dev set.  The result for ``w/o Voting'' is reported as the mean and standard deviation over multiple runs.}
\label{tab:ablation}
\begin{tabular}{lc}
\toprule
\textbf{Method} & \textbf{EX} \\
\midrule
\textbf{LEAF-SQL} & \textbf{70.9}  \\
\midrule
w/o Diversity (Single-Path Search)  & 66.2 (-4.7) \\
w/o Pruning (Brute-force Expansion)  & 68.1 (-2.8)\\
w/o Voting (Random Selection) & 67.8 (-3.1) $\pm$ 1.3  \\
\bottomrule
\end{tabular}
\end{table}

The results in Table~\ref{tab:ablation} clearly demonstrate that all components are crucial for LEAF-SQL's performance. The most significant performance degradation occurs when search diversity is removed (w/o Diversity), causing a sharp drop of 4.7 points in accuracy. This finding strongly supports our central hypothesis that exploring multiple structural hypotheses is essential to counteract the risk of early, difficult-to-recover errors in the search process.
Furthermore, removing the pruning mechanism (w/o Pruning) also leads to a performance decline of 2.8 points. This confirms the necessity of the SkeEva. Without it, low-quality skeletons propagate through the search space, introducing noise that ultimately hinders the generation of correct SQL.
Finally, replacing the voting mechanism with random selection (w/o Voting) also impairs performance. As this variant involves random sampling, we report its average performance over multiple runs, which shows a decline of 3.1 points on average. This highlights the importance of a robust result selection step to reliably identify the most plausible query from the candidate pool.

\subsection{Analysis of the Skeleton Candidate Set}
To further understand the internal mechanics of LEAF-SQL, we analyze the properties of its core output: the skeleton candidate set. This analysis provides quantitative evidence for two key characteristics of our framework: structural diversity and granularity-adaptivity. 

\subsubsection{Structural Diversity}
To analyze the structural diversity of LEAF-SQL, we conduct two experiments. First, we use the Pass@k metric to evaluate the upper-bound recall of our candidate set, demonstrating the potential of diverse generation. Second, we analyze the number of generated skeleton candidates.

We evaluate LEAF-SQL with Pass@k, which measures the fraction of problems where at least one correct SQL query is found in the candidate set. As shown in Table~\ref{tab:results_passk}, Pass@k accuracy is consistently higher than standard execution accuracy across all models. For instance, with Gemini-2.5-Pro on BIRD, accuracy rises from 73.5\% to 76.8\%. 
This result confirms that our search frequently captures the correct solution within its candidate set, highlighting the high quality and strong recall of our generation process.
As an additional reference, under the same backbone (Qwen-2.5-Coder-14B) and an identical downstream pipeline (SQL generation and voting-based selection), a simple sampling baseline with $16$ candidates achieves 61.3\% EX, compared to 67.9\% for LEAF-SQL, indicating that the gain is not solely explained by sampling more candidates.

\begin{table}[ht]
\centering
\caption{Comparison of standard execution accuracy and Pass@k accuracy. The consistent improvement in Pass@k demonstrates the value of diverse skeleton candidates.}
\label{tab:results_passk}
\resizebox{\linewidth}{!}{
    \begin{tabular}{lccc}
    \toprule[1pt]
    \textbf{Method} & \textbf{Model} & \textbf{BIRD} & \textbf{Spider} \\
    \midrule[0.5pt]
    \multirow{3}{*}{\centering LEAF-SQL} 
                    & Qwen2.5-Coder-14B & 67.9 & 86.2 \\
                    & GPT-4o & 70.9 & 88.5 \\
                    & Gemini2.5-pro & 73.5 & 89.7 \\
    \cmidrule(lr){2-4}
    \multirow{3}{*}{\centering LEAF-SQL (Pass@k)} 
                    & Qwen2.5-Coder-14B & 70.2 (+2.3) & 88.3 (+2.1) \\
                    & GPT-4o & 74.4 (+3.5) & 90.1 (+1.6) \\
                    & Gemini2.5-pro & 76.8 (+3.3) & 92.0 (+2.3) \\    
    \bottomrule[1pt]
    \end{tabular}
}
\end{table}

Next, we analyze the number of generated candidates to understand the characteristics of our search process. We vary the hyperparameter $m$ for the SkeFor, which defines the maximum number of branches explored at each step of the skeleton search. Table~\ref{tab:ablation_m_difficulty} presents the execution accuracy and the average number of skeletons searched, broken down by question difficulty.
The results reveal a key insight. The number of candidates increases with problem complexity. With our default setting ($m=3$), the framework generates only 1.4 skeletons for simple questions but increases this to 2.8 for challenging ones. This shows that LEAF-SQL intelligently allocates more search effort where it is most needed. This demonstrates that our method achieves diversity through a focused exploration of high-value alternatives rather than brute force.

\begin{table}[t]
    \centering
    \caption{Execution accuracy and the average number of skeletons generated versus max branches ($m$) and difficulty on the BIRD dev set (using GPT-4o).}
    \label{tab:ablation_m_difficulty}
    \resizebox{1\linewidth}{!}{
    \begin{tabular}{ccccc}
    \toprule[1pt]
    \multirow{2}{*}{\textbf{Max Branches ($m$)}} & \multicolumn{3}{c}{\textbf{Difficulty}} & \multirow{2}{*}{\textbf{EX}} \\
    \cmidrule(lr){2-4} 
     & \textbf{Simple} & \textbf{Moderate} & \textbf{Challenging} \\
    \midrule[0.5pt]
    1 (Greedy Search) & 1 & 1 & 1 & 66.2 \\
    2 & 1.3 & 1.8 & 2.5 & 68.7  \\
    3 (Default) & 1.4 & 2.3 & 2.8 & 70.9  \\
    \bottomrule[1pt]
    \end{tabular}
    }
\end{table}

In summary, a core idea of LEAF-SQL is to explore structurally diverse skeletons, which aligns with the real-world scenario where a single question can have multiple correct SQL solutions. 
Our analysis confirms that the generated skeleton candidates have high potential, as demonstrated by the strong Pass@k results. Furthermore, the analysis of skeleton quantities provides a more concrete explanation of this structural diversity: the number of solutions increases with problem difficulty, yet the overall count remains controlled and modest.

\subsubsection{Skeleton Granularity Distribution}

Another core idea of LEAF-SQL is its ability to adaptively adjust the granularity of the generated skeleton. As discussed in Section~\ref{sec:Analysis of Granularity Potential and Realization Difficulty}, fine-grained skeletons (e.g., Detailed) offer higher performance potential but are more challenging to generate correctly. The purpose of our adaptive mechanism is to fall back to more reliable, coarse-grained skeletons when predicting a fine-grained one is deemed too risky.

\begin{figure}[ht]
    \centering
    \includegraphics[width=1\linewidth]{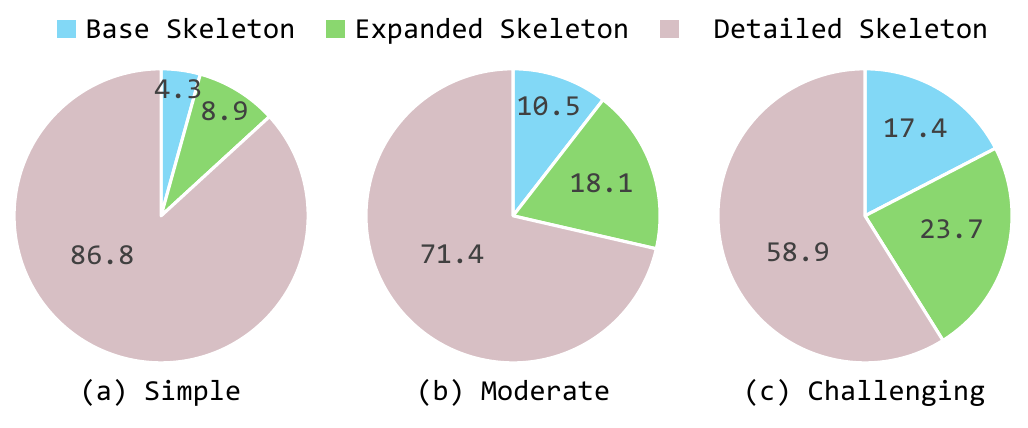} 
    \caption{Granularity distribution of skeletons generated by LEAF-SQL across Simple, Moderate, and Challenging subsets of the BIRD benchmark.}
    \label{fig:granularity_distribution}
\end{figure}

To demonstrate this adaptive behavior, we analyze the granularity distribution of the skeletons yielded by LEAF-SQL across different task difficulties on the BIRD benchmark. Figure~\ref{fig:granularity_distribution} shows the proportion of Base, Expanded, and Detailed skeletons for simple, moderate, and challenging questions.

The results clearly illustrate LEAF-SQL's adaptive strategy. For simple questions, the framework confidently generates fine-grained skeletons, with the Detailed type accounting for a significant 86.8\% share, aiming for maximum performance. However, as the problem difficulty increases, LEAF-SQL becomes progressively more conservative. For challenging questions, the proportion of the safest, coarse-grained Base skeletons rises sharply from 8.9\% to 23.7\%, while the share of the riskiest Detailed skeletons drops from 86.8\% to 58.9\%. This demonstrates that LEAF-SQL intelligently trades off the high potential of fine-grained skeletons against the high reliability of coarse-grained ones, effectively balancing performance and robustness based on task complexity.

\subsection{Token and Time Cost}
\label{sec:Token Cost}
\begin{figure}[ht]
    \centering
    \includegraphics[width=1\linewidth]{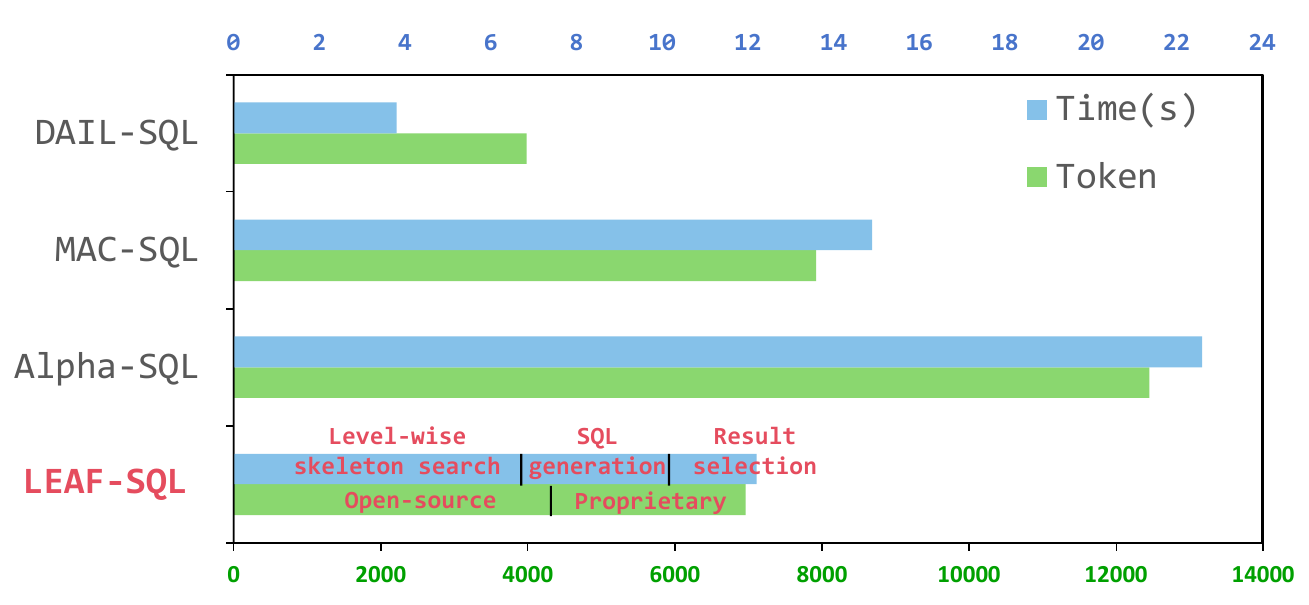}
    \caption{Average consumption of time and tokens on BIRD.}
    \label{fig:analysis_cost}
\end{figure}

Although LEAF-SQL introduces additional overheads, it remains reasonably efficient. Figure~\ref{fig:analysis_cost} reports average execution time and token consumption on 100 randomly sampled examples from BIRD.

\begin{figure*}[ht]
    \centering
    \includegraphics[width=1\linewidth]{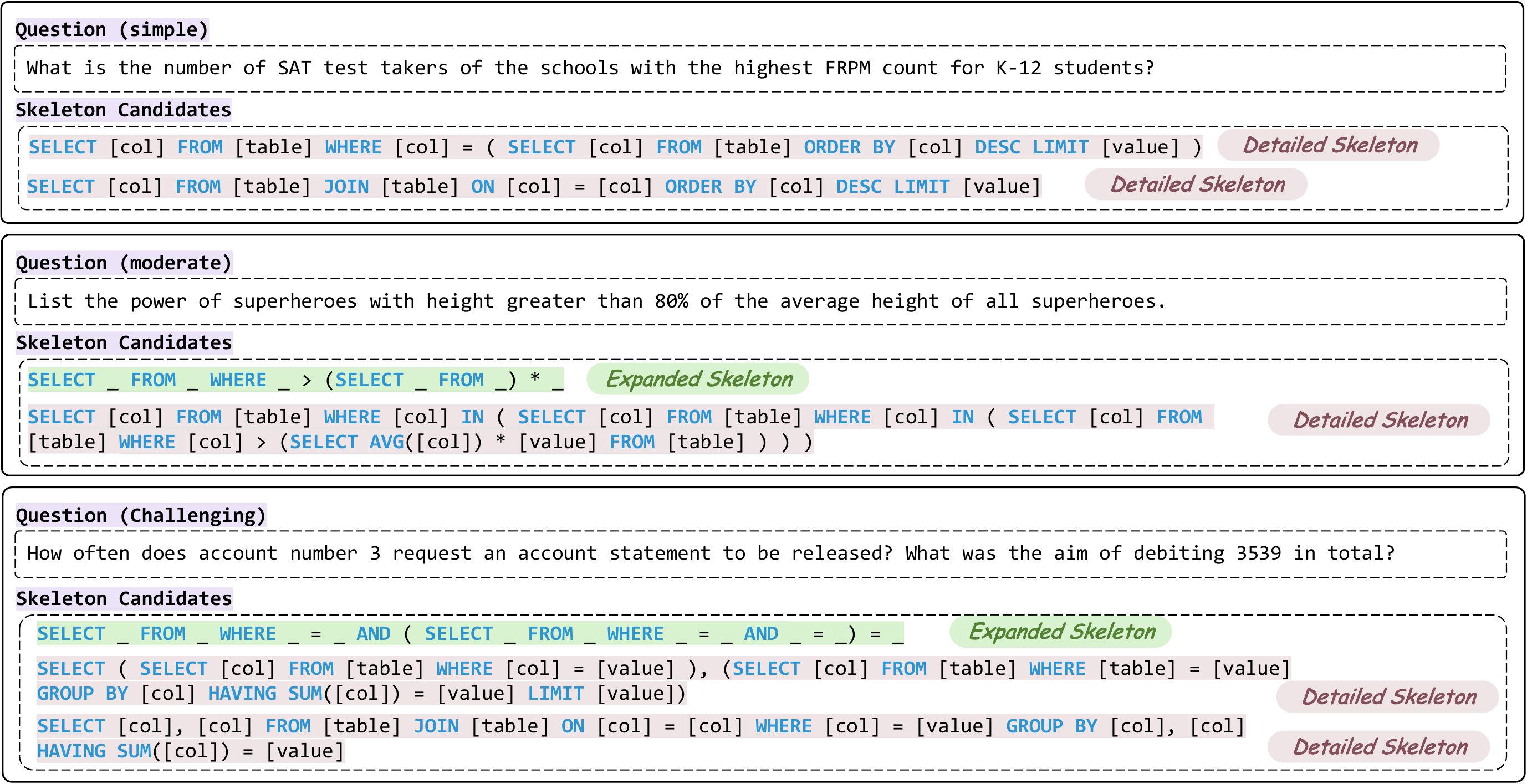}
    \caption{Case studies from the BIRD benchmark illustrating the outputs of LEAF-SQL. For each question of varying difficulty (simple, moderate, challenging), our method generates a diverse set of skeleton candidates with adaptive granularity.}
    \label{fig:case study}
\end{figure*}

By implementing Asynchronous Streaming, we decouple skeleton generation and evaluation via asynchronous queues to improve throughput under concurrent load. This reduces the amortized per-query time of the LWSS module to 6.2 seconds. The end-to-end latency is controlled within $\sim$12 seconds, making LEAF-SQL practically competitive in deployment.
To manage computational costs, we employ Demonstration Pruning: in the few-shot demonstrations, we retain only the query-relevant ``Gold Schema''. Importantly, this pruning is applied only to demonstrations constructed offline; no gold schema is used at test time.

\subsection{Case Study}
\label{sec:Case Study}
Figure~\ref{fig:case study} provides intuitive examples of LEAF-SQL's mechanics. For the \emph{simple} question, it explores structural diversity by generating two distinct Detailed skeletons (a subquery vs. a \texttt{JOIN}). For the \emph{moderate} question, it demonstrates granularity adaptivity by proposing both a safe Expanded skeleton and an ambitious Detailed one. Finally, for the \emph{challenging} question, the framework combines both strategies, generating a versatile candidate set comprising one Expanded and two distinct Detailed skeletons to maximize the chance of success.






\subsection{Discussion and Limitations}
While LEAF-SQL demonstrates strong performance, we acknowledge several limitations regarding its practical deployment and robustness. 
First, although parallelization improves amortized throughput, the reliance on interactions with multiple LLMs still incurs high single-query latency, making the framework less suitable for strictly real-time applications. 
Second, the branching and pruning mechanisms in SkeFor and SkeEva rely on heuristic prompting. Although effective on current benchmarks, their stability under real-world queries requires further validation. 
Finally, while our method excels at handling complex logic on standard benchmarks, it has not been stress-tested on large enterprise schemas. 
Therefore, we currently position LEAF-SQL as an exploratory framework for Text-to-SQL tasks, rather than a fully optimized solution for large-scale or real-time enterprise deployments.

\section{Conclusion}
In this paper, we introduce LEAF-SQL, a novel framework that overcomes key limitations in existing skeleton-based Text-to-SQL methods. By reframing skeleton prediction as a coarse-to-fine tree search, LEAF-SQL efficiently navigates a broad solution space, yielding a set of candidates that is both structurally diverse and granularity-adaptive. LEAF-SQL is driven by several key techniques: a three-level skeleton hierarchy to guide the search, a Skeleton Formulation Agent to generate diverse hypotheses, and a Skeleton Evaluation Agent to refine the process through intelligent pruning.
Our extensive experiments and in-depth analyses validate the effectiveness of this approach. LEAF-SQL consistently improves the performance of various LLM backbones. On the official hidden BIRD test set leaderboard, LEAF-SQL achieves a SOTA execution accuracy of 71.6\%, surpassing the leading search-based method, Alpha-SQL, by 1.4 points. Furthermore, LEAF-SQL achieves 89.7\% on the Spider dev set and 73.5\% on the BIRD dev set, outperforming the current top skeleton-based approach UCS-SQL by 2.4 and 4.5 points respectively.
We have demonstrated that our framework's strength lies in its ability to explore multiple valid logical structures and dynamically balance performance with robustness by adapting skeleton granularity based on task complexity. These findings confirm that a search-based exploration of skeletons provides a more robust and powerful foundation for tackling complex Text-to-SQL tasks.

\section*{Acknowledgment}
This work was supported by the National Natural Science Foundation of China (Grant Nos. 62462034, 62562033, 62272205, 62272206), the Natural Science Foundation of Jiangxi Province (Grant Nos. 20232ACB202008, 20242BAB25119), and the Jiangxi Provincial Graduate Innovation Special Fund Project (Grant No. YC2023-B185).

\section*{AI-Generated Content Acknowledgement}
We lightly used OpenAI GPT and Google Gemini to improve readability and perform grammar checks on author-written text, and to suggest small code snippets for non-core utilities. All scientific ideas, core algorithms and implementations (including the LEAF-SQL skeleton hierarchy, agents, and search), experimental setup and results are entirely authored, verified, and validated by the authors; no AI-generated content was used to produce or substantiate the technical contributions.

\bibliographystyle{IEEEtran}
\bibliography{leaf_sql}

@inproceedings{wu-etal-2025-ucs,
    title = {UCS-SQL: Uniting Content and Structure for Enhanced Semantic Bridging In Text-to-SQL},
    author = "Wu, Zhenhe  and
      Li, Zhongqiu  and
      JieZhangChinaTele, JieZhangChinaTele  and
      He, Zhongjiang  and
      Yang, Jian  and
      Zhao, Yu  and
      Fang, Ruiyu  and
      Wang, Bing  and
      Xie, Hongyan  and
      Song, Shuangyong  and
      Li, Zhoujun",
    booktitle = "Findings of the Association for Computational Linguistics: ACL 2025",
    year = "2025",
    address = "Vienna, Austria",
    publisher = "Association for Computational Linguistics",
    pages = "8156--8168",
}

@inproceedings{yu-etal-2018-spider,
    title = {Spider: A Large-Scale Human-Labeled Dataset for Complex and Cross-Domain Semantic Parsing and Text-to-SQL Task},
    author = "Yu, Tao  and
      Zhang, Rui  and
      Yang, Kai  and
      Yasunaga, Michihiro  and
      Wang, Dongxu  and
      Li, Zifan  and
      Ma, James  and
      Li, Irene  and
      Yao, Qingning  and
      Roman, Shanelle  and
      Zhang, Zilin  and
      Radev, Dragomir",
    editor = "",
    booktitle = "Proceedings of the 2018 Conference on Empirical Methods in Natural Language Processing",
    month = "",
    year = "2018",
    address = "Brussels, Belgium",
    publisher = "Association for Computational Linguistics",
    url = "",
    doi = "",
    pages = "3911--3921"
}

@inproceedings{li2023bird,
  author       = {Jinyang Li and Binyuan Hui and Ge Qu and Jiaxi Yang and Binhua Li and Bowen Li and Bailin Wang and Bowen Qin and Ruiying Geng and Nan Huo and Xuanhe Zhou and Chenhao Ma and Guoliang Li and Kevin C.C. Chang and Fei Huang and Reynold Cheng and Yongbin Li},
  title        = {Can LLM Already Serve as a Database Interface? A Big Bench for Large-Scale Database Grounded Text-to-SQLs},
  booktitle    = {Proceedings of the 37th International Conference on Neural Information Processing Systems (NeurIPS)},
  year         = {2023},
  publisher    = {Curran Associates Inc.},
  address      = {Red Hook, NY, USA},
  pages        = {1835:1--1835:28},
  url          = {},
}

@article{dail-sql,
author = {Gao, Dawei and Wang, Haibin and Li, Yaliang and Sun, Xiuyu and Qian, Yichen and Ding, Bolin and Zhou, Jingren},
title = {Text-to-SQL Empowered by Large Language Models: A Benchmark Evaluation},
year = {2024},
issue_date = {January 2024},
publisher = {VLDB Endowment},
volume = {17},
number = {5},
issn = {2150-8097},
journal = {Proc. VLDB Endow.},
pages = {1132–1145},
numpages = {14}
}

@article{rsl-sql,
      title={RSL-SQL: Robust Schema Linking in Text-to-SQL Generation}, 
      author={Zhenbiao Cao and Yuanlei Zheng and Zhihao Fan and Xiaojin Zhang and Wei Chen and Xiang Bai},
      year={2024},
      journal       = {arXiv preprint arXiv:2411.00073},
}

@article{alpha-sql,
      title={Alpha-SQL: Zero-Shot Text-to-SQL using Monte Carlo Tree Search}, 
      author={Boyan Li and Jiayi Zhang and Ju Fan and Yanwei Xu and Chong Chen and Nan Tang and Yuyu Luo},
      year={2025},
      journal       = {arXiv preprint arXiv:2502.17248},

}

@article{yuan2025mctssqllightweightllmsmaster,
      title=  {MCTS-SQL: Light-Weight LLMs can Master the Text-to-SQL through Monte Carlo Tree Search}, 
      author={Shuozhi Yuan and Limin Chen and Miaomiao Yuan and Jin Zhao},
      year={2025},
      journal       = {arXiv preprint arXiv:2501.16607},
}

@inproceedings{mac-sql,
    title = {MAC-SQL: A Multi-Agent Collaborative Framework for Text-to-SQL},
    author = "Wang, Bing  and
      Ren, Changyu  and
      Yang, Jian  and
      Liang, Xinnian  and
      Bai, Jiaqi  and
      Chai, LinZheng  and
      Yan, Zhao  and
      Zhang, Qian-Wen  and
      Yin, Di  and
      Sun, Xing  and
      Li, Zhoujun",

    booktitle = "Proceedings of the 31st International Conference on Computational Linguistics",

    year = "2025",
    address = "Abu Dhabi, UAE",

    pages = "540--557"
}

@article{supersql,
author = {Li, Boyan and Luo, Yuyu and Chai, Chengliang and Li, Guoliang and Tang, Nan},
title = {The Dawn of Natural Language to SQL: Are We Fully Ready?},
year = {2024},
issue_date = {July 2024},
publisher = {VLDB Endowment},
volume = {17},
number = {11},
issn = {2150-8097},
journal = {Proc. VLDB Endow.},
pages = {3318–3331},
numpages = {14}
}

@article{learnat,
        title={LearNAT: Learning NL2SQL with AST-guided Task Decomposition for Large Language Models}, 
      author={Weibin Liao and Xin Gao and Tianyu Jia and Rihong Qiu and Yifan Zhu and Yang Lin and Xu Chu and Junfeng Zhao and Yasha Wang},
      year={2025},
      journal       = {arXiv preprint arXiv:2504.02327},
      url={}, 
}

@article{10.1145/3589292,
author = {Gu, Zihui and Fan, Ju and Tang, Nan and Cao, Lei and Jia, Bowen and Madden, Sam and Du, Xiaoyong},
title = {Few-shot Text-to-SQL Translation using Structure and Content Prompt Learning},
year = {2023},
publisher = {Association for Computing Machinery},
address = {New York, NY, USA},
volume = {1},
pages        = {138--166},
number = {2},
url = {},
doi = {},
journal = {Proc. ACM Manag. Data},
articleno = {147},
numpages = {28}
}

@inproceedings{li2023resdsql,
  title        = {RESDSQL: Decoupling Schema Linking and Skeleton Parsing for Text-to-SQL},
  author       = {Li, Hao and Zhang, Jichuan and Li, Cheng and Chen, Hong},
  booktitle    = {Proceedings of the AAAI Conference on Artificial Intelligence},
  volume       = {37},
  number       = {11},
  pages        = {13067--13075},
  year         = {2023},
}

@article{10.14778/3681954.3681960,
author = {Fan, Ju and Gu, Zihui and Zhang, Songyue and Zhang, Yuxin and Chen, Zui and Cao, Lei and Li, Guoliang and Madden, Samuel and Du, Xiaoyong and Tang, Nan},
title = {Combining Small Language Models and Large Language Models for Zero-Shot NL2SQL},
year = {2024},
publisher = {VLDB Endowment},
volume = {17},
number = {11},
issn = {2150-8097},

journal = {Proc. VLDB Endow.},
pages = {2750–2763},
numpages = {14}
}

@ARTICLE{11095853,
  author={Liu, Xinyu and Shen, Shuyu and Li, Boyan and Ma, Peixian and Jiang, Runzhi and Zhang, Yuxin and Fan, Ju and Li, Guoliang and Tang, Nan and Luo, Yuyu},
  journal={IEEE Transactions on Knowledge and Data Engineering}, 
  title={A Survey of Text-to-SQL in the Era of LLMs: Where Are We, and Where Are We Going?}, 
  year={2025},
  volume={37},
  number={10},
  pages={5735-5754},
  doi={10.1109/TKDE.2025.3592032}}

@article{zhong2017seq2sqlgeneratingstructuredqueries,
      title={Seq2SQL: Generating Structured Queries from Natural Language using Reinforcement Learning}, 
      author={Victor Zhong and Caiming Xiong and Richard Socher},
      year={2017},
      journal       = {arXiv preprint arXiv:1709.00103},
}

@article{lei2025spider20evaluatinglanguage,
      title={Spider 2.0: Evaluating Language Models on Real-World Enterprise Text-to-SQL Workflows}, 
      author={Fangyu Lei and Jixuan Chen and Yuxiao Ye and Ruisheng Cao and Dongchan Shin and Hongjin Su and Zhaoqing Suo and Hongcheng Gao and Wenjing Hu and Pengcheng Yin and Victor Zhong and Caiming Xiong and Ruoxi Sun and Qian Liu and Sida Wang and Tao Yu},
      year={2025},
      journal       = {arXiv preprint arXiv:2411.07763},
}

@inproceedings{mao-etal-2024-enhancing,
    title = {Enhancing Text-to-SQL Parsing through Question Rewriting and Execution-Guided Refinement},
    author = "Mao, Wenxin  and
      Wang, Ruiqi  and
      Guo, Jiyu  and
      Zeng, Jichuan  and
      Gao, Cuiyun  and
      Han, Peiyi  and
      Liu, Chuanyi",
    booktitle = "Findings of the Association for Computational Linguistics: ACL 2024",
    year = "2024",
    address = "Bangkok, Thailand",
    publisher = "Association for Computational Linguistics",
    pages = "2009--2024"
}

@inproceedings{qu-etal-2025-share,
    title = {SHARE: An SLM-based Hierarchical Action CorREction Assistant for Text-to-SQL},
    author = "Qu, Ge  and
      Li, Jinyang  and
      Qin, Bowen  and
      Li, Xiaolong  and
      Huo, Nan  and
      Ma, Chenhao  and
      Cheng, Reynold",

    booktitle = "Proceedings of the 63rd Annual Meeting of the Association for Computational Linguistics",

    year = "2025",
    address = "Vienna, Austria",
    publisher = "Association for Computational Linguistics",
   
    pages = "11268--11292",
    ISBN = "979-8-89176-251-0"
}

@article{lou-etal-2024-large,
    title = "Large Language Model Instruction Following: A Survey of Progresses and Challenges",
    author = "Lou, Renze  and
      Zhang, Kai  and
      Yin, Wenpeng",
    journal = "Computational Linguistics",
    volume = "50",
    number = "3",
    year = "2024",
    address = "Cambridge, MA",
    publisher = "MIT Press",
    pages = "1053--1095"
}

@inproceedings{tai-etal-2023-exploring,
    title = {Exploring Chain of Thought Style Prompting for Text-to-SQL},
    author = "Tai, Chang-Yu  and
      Chen, Ziru  and
      Zhang, Tianshu  and
      Deng, Xiang  and
      Sun, Huan",

    booktitle = "Proceedings of the 2023 Conference on Empirical Methods in Natural Language Processing",

    year = "2023",
    address = "Singapore",
    publisher = "Association for Computational Linguistics",

    pages = "5376--5393"
}

@article{chen2023teachinglargelanguagemodels,
      title={Teaching Large Language Models to Self-Debug}, 
      author={Xinyun Chen and Maxwell Lin and Nathanael Schärli and Denny Zhou},
      year={2023},
      journal       = {arXiv preprint arXiv:2304.05128},
}

@article{wang2024dacdecomposedautomationcorrection,
      title={DAC: Decomposed Automation Correction for Text-to-SQL}, 
      author={Dingzirui Wang and Longxu Dou and Xuanliang Zhang and Qingfu Zhu and Wanxiang Che},
      year={2024},
      journal       = {arXiv preprint arXiv:2408.08779},
}

@inproceedings{wu-etal-2023-chain,
    title = "Chain of Thought Prompting Elicits Knowledge Augmentation",
    author = "Wu, Dingjun  and
      Zhang, Jing  and
      Huang, Xinmei",
    booktitle = "Findings of the Association for Computational Linguistics: ACL 2023",
    
    address = "Toronto, Canada",
    publisher = "Association for Computational Linguistics",
    pages = "6519--6534",
    year = "2023",
}

@article{liu2025xiyansqlnovelmultigeneratorframework,
      title={XiYan-SQL: A Novel Multi-Generator Framework For Text-to-SQL}, 
      author={Yifu Liu and Yin Zhu and Yingqi Gao and Zhiling Luo and Xiaoxia Li and Xiaorong Shi and Yuntao Hong and Jinyang Gao and Yu Li and Bolin Ding and Jingren Zhou},
      year={2025},
      journal       = {arXiv preprint arXiv:2507.04701},
}

@inproceedings{lee-etal-2025-dcg,
    title = {DCG-SQL: Enhancing In-Context Learning for Text-to-SQL with Deep Contextual Schema Link Graph},
    author = "Lee, Jihyung  and
      Lee, Jin-Seop  and
      Lee, Jaehoon  and
      Choi, YunSeok  and
      Lee, Jee-Hyong",
    
    booktitle = "Proceedings of the 63rd Annual Meeting of the Association for Computational Linguistics",
    year = "2025",
    address = "Vienna, Austria",

    pages = "15397--15412",
}

@inproceedings{dai-etal-2025-parsql,
    title = {PARSQL: Enhancing Text-to-SQL through SQL Parsing and Reasoning},
    author = "Dai, Yaxun  and
      Yang, Haiqin  and
      Hao, Mou  and
      Chao, Pingfu",

    booktitle = "Findings of the Association for Computational Linguistics: ACL 2025",

    year = "2025",
    address = "Vienna, Austria",
 
    pages = "661--681",

}

@inproceedings{yang-etal-2024-synthesizing,
    title = {Synthesizing Text-to-SQL Data from Weak and Strong LLMs},
    author = "Yang, Jiaxi  and
      Hui, Binyuan  and
      Yang, Min  and
      Yang, Jian  and
      Lin, Junyang  and
      Zhou, Chang",

    booktitle = "Proceedings of the 62nd Annual Meeting of the Association for Computational Linguistics",
    year = "2024",
    address = "Bangkok, Thailand",
    pages = "7864--7875",
  
}

@inproceedings{gan-etal-2021-towards,
    title = {Towards Robustness of Text-to-SQL Models against Synonym Substitution},
    author = "Gan, Yujian  and
      Chen, Xinyun  and
      Huang, Qiuping  and
      Purver, Matthew  and
      Woodward, John R.  and
      Xie, Jinxia  and
      Huang, Pengsheng",

    booktitle = "Proceedings of the 59th Annual Meeting of the Association for Computational Linguistics and the 11th International Joint Conference on Natural Language Processing",

    year = "2021",
    address = "Online",
   
    pages = "2505--2515",
}

@inproceedings{deng-etal-2021-structure,
    title = {Structure-Grounded Pretraining for Text-to-SQL},
    author = "Deng, Xiang  and
      Awadallah, Ahmed Hassan  and
      Meek, Christopher  and
      Polozov, Oleksandr  and
      Sun, Huan  and
      Richardson, Matthew",

    booktitle = "Proceedings of the 2021 Conference of the North American Chapter of the Association for Computational Linguistics: Human Language Technologies",

    year = "2021",
    address = "Online",
    

    pages = "1337--1350",
   
}

@inproceedings{gan-etal-2021-exploring,
    title = {Exploring Underexplored Limitations of Cross-Domain Text-to-SQL Generalization},
    author = "Gan, Yujian  and
      Chen, Xinyun  and
      Purver, Matthew",

    booktitle = "Proceedings of the 2021 Conference on Empirical Methods in Natural Language Processing",

    year = "2021",
    address = "Online and Punta Cana, Dominican Republic",
    

    pages = "8926--8931",
   
}

@article{yang2025qwen3technicalreport,
      title={Qwen3 Technical Report}, 
      author={An Yang and Anfeng Li and Baosong Yang and Beichen Zhang and others},
      year={2025},
      journal       = {arXiv preprint arXiv:2505.09388},
}

@article{openai2024gpt4technicalreport,
      title={GPT-4 Technical Report}, 
      author={OpenAI and Josh Achiam and Steven Adler and Sandhini Agarwal and Lama Ahmad and others},
      year={2024},
      journal       = {arXiv preprint arXiv:2303.08774},
}

@article{comanici2025gemini25pushingfrontier,
      title={Gemini 2.5: Pushing the Frontier with Advanced Reasoning, Multimodality, Long Context, and Next Generation Agentic Capabilities}, 
      author={Gheorghe Comanici and Eric Bieber and Mike Schaekermann and Ice Pasupat and Noveen Sachdeva and others},
      year={2025},
      journal       = {arXiv preprint arXiv:2507.06261},
}

@article{qwen2025qwen25technicalreport,
      title={Qwen2.5 Technical Report}, 
      author={An Yang and Baosong Yang and Beichen Zhang and Binyuan Hui and Bo Zheng and others},
      year={2024},
      journal       = {arXiv preprint arXiv:2412.15115},
}

@inproceedings{10.5555/3666122.3666639,
author = {Yao, Shunyu and Yu, Dian and Zhao, Jeffrey and Shafran, Izhak and Griffiths, Thomas L. and Cao, Yuan and Narasimhan, Karthik},
title = {Tree of thoughts: deliberate problem solving with large language models},
year = {2023},
address = {Red Hook, NY, USA},
articleno = {517},
numpages = {14},
pages = {8812-8825},
booktitle = {Proceedings of the 37th International Conference on Neural Information Processing Systems},
articleno = {517},
numpages = {14},
}

@article{lyu2025sqlo1selfrewardheuristicdynamic,
      title={SQL-o1: A Self-Reward Heuristic Dynamic Search Method for Text-to-SQL}, 
      author={Shuai Lyu and Haoran Luo and Ripeng Li and Zhonghong Ou and Jiangfeng Sun and Yang Qin and Xiaoran Shang and Meina Song and Yifan Zhu},
      year={2025},
      journal       = {arXiv preprint arXiv:2502.11741},
      url={}, 
}

@article{ren2024purplemakinglargelanguage,
      title={PURPLE: Making a Large Language Model a Better SQL Writer}, 
      author={Tonghui Ren and Yuankai Fan and Zhenying He and Ren Huang and Jiaqi Dai and Can Huang and Yinan Jing and Kai Zhang and Yifan Yang and X. Sean Wang},
      year={2024},
      eprint={2403.20014},
      journal       = {arXiv preprint arXiv:2403.20014},
}

@article{csc-sql,
      title={CSC-SQL: Corrective Self-Consistency in Text-to-SQL via Reinforcement Learning}, 
      author={Lei Sheng and Shuai-Shuai Xu},
      year={2025},
      eprint={2505.13271},
      journal       = {arXiv preprint arXiv:2505.13271},
}

\end{document}